%% file: neurips_2026.tex
\documentclass{article}

\PassOptionsToPackage{numbers, compress}{natbib}
\usepackage[preprint]{neurips_2026}

\usepackage[utf8]{inputenc}
\usepackage[T1]{fontenc}
\usepackage{amsmath}
\usepackage{amssymb}
\usepackage{amsfonts}
\usepackage{booktabs}
\usepackage{graphicx}
\usepackage{multirow}
\usepackage[table]{xcolor}
\usepackage{nicefrac}
\usepackage{microtype}
\usepackage{url}
\usepackage{algorithm}
\usepackage{algpseudocode}
\usepackage{caption}
\usepackage{subcaption}
\usepackage{tcolorbox}
\tcbuselibrary{skins, breakable}
\usepackage{enumitem}
\usepackage{wrapfig}
\usepackage[accsupp]{axessibility}
\usepackage[hidelinks]{hyperref}

\newcounter{equalcontribcount}
\newcommand{\equalcontrib}{%
  \stepcounter{equalcontribcount}%
  \ifnum\value{equalcontribcount}=1
    \thanks{These authors contributed equally.}%
  \else
    \footnotemark[1]%
  \fi
}

\makeatletter
\newcommand\blfootnote[1]{%
  \begingroup
  \renewcommand\thefootnote{}\footnote{#1}%
  \addtocounter{footnote}{-1}%
  \endgroup
}
\makeatother

\newcommand{\authorskip}{\hspace{1.5mm}}

\definecolor{codebasecolor}{HTML}{2A6F97}

\title{DynFrame: Adaptive Reasoning-Driven Multimodal \\
  Framework with Dynamic Frame Augmentation \\
  for Complex Video Understanding}

\author{%
  \textbf{Peng Zhang\textsuperscript{1,2,*}\authorskip
  Guanghao Zhang\textsuperscript{2,*,\S}\authorskip
  Wanggui He\textsuperscript{2}\authorskip
  Longxiang Zhang\textsuperscript{2}\authorskip
  Mushui Liu\textsuperscript{1,2}\authorskip
  Yan Xia\textsuperscript{2}} \\
  \textbf{Zhenhao Peng\textsuperscript{2}\authorskip
  Weilong Dai\textsuperscript{2}\authorskip
  Jinlong Liu\textsuperscript{2}\authorskip
  Haobing Tang\textsuperscript{2}\authorskip
  Le Zhang\textsuperscript{2}\authorskip
  Hao Jiang\textsuperscript{2,\dag}\authorskip
  Pipei Huang\textsuperscript{2}\authorskip} \\[0.5em]
  {\normalfont \textsuperscript{1}Zhejiang University\quad
  \textsuperscript{2}Alibaba Group}
}

\begin{document}
\maketitle

\blfootnote{\hspace*{-1.8em}\textsuperscript{*}Equal contribution.\quad
\textsuperscript{\S}Project lead.\quad
\textsuperscript{\dag}Corresponding author.}

\renewcommand{\topfraction}{0.92}
\renewcommand{\bottomfraction}{0.30}
\renewcommand{\textfraction}{0.06}
\renewcommand{\floatpagefraction}{0.85}
\renewcommand{\dbltopfraction}{0.92}
\renewcommand{\dblfloatpagefraction}{0.85}
\setcounter{topnumber}{3}
\setcounter{bottomnumber}{2}
\setcounter{totalnumber}{5}
\setcounter{dbltopnumber}{3}
\setlength{\textfloatsep}{10pt plus 2pt minus 2pt}
\setlength{\floatsep}{8pt plus 2pt minus 2pt}
\setlength{\intextsep}{8pt plus 2pt minus 2pt}

\input{sections/00_abstract}

\input{sections/01_introduction}

\input{sections/02_related_work}
\input{sections/03_method}
\input{sections/04_experiments}

\input{sections/07_conclusion}

\clearpage
\bibliographystyle{plainnat}
{\small
\bibliography{main}
}

\clearpage
\input{appendix}

\end{document}

%% file: sections/00_abstract.tex
\begin{abstract}
Recent video multimodal large language models (MLLMs) increasingly
couple step-by-step reasoning with on-demand visual evidence retrieval,
allowing models to revisit relevant video segments during inference.
However, two structural gaps remain in existing thinking-with-video
systems. \textbf{(i)} Sampling density is not a learnable decision:
existing methods may let the model decide \emph{where} to look, but the
per-window frame rate is largely fixed. As a result, fine-grained evidence is often recovered
through repeated retrieval calls which
increases inference context length and training difficulty.  \textbf{(ii)} Retrieval and answer generation are
usually optimized with a single trajectory-level advantage, so the ``where to
look'' tokens and the ``how to answer'' tokens receive the same credit
even when one is correct and the other is not.
To address these gaps, we present \textbf{DynFrame}, a framework that emits the temporal window
and the sampling density as native tokens within a single
autoregressive pass, such learnable span--density retrieval enables
acquiring multi-granularity evidence with a single retrieval step. Based on the above tokenized retrieval interface, we further introduce
\textbf{Segment-Decoupled GRPO (SD-GRPO)}, which splits each rollout at
the retrieval boundary and assigns role-specific token-level advantages, separately crediting the
sampling decision and the answer.
Trained on the curated \textbf{DM-CoT-74k} and \textbf{DM-RL-45k}, DynFrame-4B is competitive with strong 7B--8B baselines across six
benchmarks (NExT-GQA, Charades-STA, ActivityNet-MR, Video-MME, MLVU,
LVBench), and DynFrame-8B sets new state-of-the-art on most metrics.
Code is available at {\hypersetup{urlcolor=codebasecolor}\href{https://github.com/zhangguanghao523/DynFrame}{https://github.com/zhangguanghao523/DynFrame}}.
\end{abstract}

%% file: sections/01_introduction.tex
\section{Introduction}
\label{sec:intro}

Multimodal large language models (MLLMs)~\cite{
bai2025qwen2.5vl,zhu2025internvl3,li2024llavaonevision,
hurst2024gpt4o,comanici2025gemini25,kimi2025vl,bai2025qwen3vl} have
substantially advanced video question answering, temporal grounding, and
long-form video understanding~\cite{
nextgqa,gao2017tall,song2024temporalgrounding,krishna2017dense,
fu2025videomme,mlvu,wang2024lvbench}.
Building on chain-of-thought (CoT)
reasoning~\cite{wei2022chain,shao2024visual,wang2025multimodal},
recent supervised and reinforcement-learning
post-training methods further improve multi-step video inference~\cite{
feng2025videor1,li2025videochatr1,wang2025videorft}.
However, most video reasoners still operate after a fixed visual pass:
a sparse set of frames is selected before generation, and all subsequent
intermediate steps are purely textual~\cite{
maaz2023video,zhang2023videollama,jin2024chat,zhang2024video,
bai2025qwen3vl}. This design is fragile for long
videos, where the answer may hinge on a short action, a brief object
state change, or multiple clues scattered across a redundant temporal
context. As reasoning chains grow, the model can drift away
from the actual visual evidence and hallucinate missing fine-grained
events.

\begin{figure}[t]
    \centering
    \includegraphics[width=\linewidth]{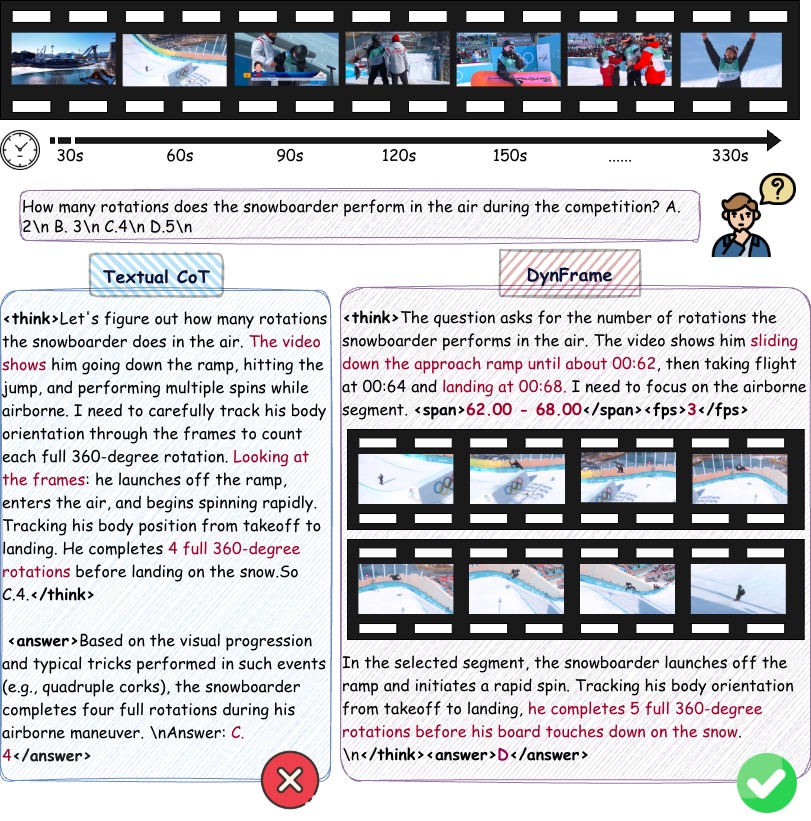}
    \caption{\textbf{Textual CoT vs. DynFrame.}
    Textual CoT (left) reasons over a fixed sparse frame set and
    misses the airborne segment, yielding a wrong rotation count.
    DynFrame (right) emits \texttt{<span>} and \texttt{<fps>}
    tokens \emph{within} its reasoning to retrieve a denser,
    temporally focused frame set, and then continues reasoning over
    the augmented visual context to reach the correct answer.}
    \label{fig:teaser}
\end{figure}

A growing \emph{thinking-with-video} line of work addresses this
limitation by allowing models to actively revisit video evidence during
inference. Existing systems instantiate visual revisiting through
several retrieval interfaces: tool-based clip retrieval that calls an
external module to crop or resample candidate video segments
~\cite{vital2025,yang2025longvt}, zoom-in or temporal-focusing
mechanisms that inspect local regions at higher resolution
~\cite{lover12025,ding2025videozoomer}, multi-turn frame spotlighting
or iterative perception that progressively refines clue-focused
temporal regions~\cite{framethinker2025,videochatr152025}, and native
interleaved tool invocation that couples evidence seeking with
reasoning in a shared context~\cite{zeng2026videoo3,openo3video2025,
videotempo32026}. These systems demonstrate the importance of
revisiting visual evidence, but fine-grained evidence acquisition is
often achieved by issuing repeated retrieval calls, refining temporal
clues across turns, or appending additional high-resolution clips into
an expanding context. Such multi-turn retrieval increases inference
context length and also complicates training. This motivates a
complementary question: \emph{can a video MLLM make each retrieval
action more expressive, so that task-adaptive, multi-granularity
evidence can be acquired with fewer retrieval steps?}

To address these challenges, we introduce \textbf{DynFrame}, a novel framework that emits the temporal window and the sampling density as native tokens within a single autoregressive pass, turning frame-rate adaptation
from a system hyperparameter into a learnable per-step decision. This
learnable span--density interface enables task-adaptive frame
acquisition: the model can retrieve dense frames for short,
motion-sensitive events and sparse frames for long-range semantic
understanding, acquiring multi-granularity evidence with a single
retrieval step. This reduces reliance on repeated multi-round
retrieval calls, which often introduce long inference contexts and
complex tool-call training designs.

Furthermore, based on the explicit retrieval boundary created by this tokenized interface, we propose \textbf{Segment-Decoupled GRPO (SD-GRPO)}, which
splits each rollout at the retrieval boundary and assigns
role-specific token-level advantages, separately crediting the sampling
decision and the answer reasoning. This provides targeted credit for temporal selection and sampling
density while preserving the end-to-end answer signal. To train this behavior, we curate
task-balanced \textbf{DM-CoT-74k} for supervised fine-tuning and
\textbf{DM-RL-45k} for reinforcement learning, explicitly designed to
cultivate robust native tokenized adaptive retrieval capabilities.
Extensive experiments across six benchmarks spanning temporal
grounding, grounded VideoQA, and long-form video understanding show
that DynFrame-4B is competitive with strong 7B--8B baselines, while
DynFrame-8B achieves state-of-the-art results on most metrics.

In summary, our contributions are three-fold:
\begin{itemize}
    \item We introduce \textbf{DynFrame}, a multimodal
    reasoning framework that emits temporal span and sampling density
    as native tokens, turning adaptive temporal evidence acquisition from an externally
scheduled tool operation into a model-native reasoning capability.

    \item We curate \textbf{DM-CoT-74k} for cold-start SFT and
    \textbf{DM-RL-45k} for reinforcement learning, and propose
    \textbf{SD-GRPO}, which uses the explicit retrieval boundary to
    assign role-specific token-level advantages to the sampling and
    reasoning segments.

    \item Across six benchmarks spanning temporal grounding, grounded
    VideoQA, and long-form video understanding, DynFrame-4B is competitive with strong 7B--8B baselines, while
DynFrame-8B achieves state-of-the-art results on most metrics.
\end{itemize}

%% file: sections/02_related_work.tex
\section{Related Work}
\label{sec:related}

\paragraph{Multimodal Chain-of-Thought for video reasoning.}
Multimodal chain-of-thought extends textual CoT~\cite{wei2022chain}
by allowing intermediate reasoning to interact with visual evidence
rather than relying on a fixed visual pass. Early video reasoning and
post-training methods, such as Video-R1~\cite{feng2025videor1},
VideoChat-R1~\cite{li2025videochatr1},
VideoRFT~\cite{wang2025videorft},
Temporal-RLT~\cite{li2025temporalrlt},
improve multi-step video inference
through supervised fine-tuning or reinforcement learning, but they
mostly reason over the visual tokens supplied at the beginning of
generation. A growing \emph{thinking-with-video} line instead lets the
model actively revisit visual evidence during inference.
VITAL~\cite{vital2025} and LongVT~\cite{yang2025longvt} formulate
evidence acquisition as tool-based clip retrieval;
LOVE-R1~\cite{lover12025} and
VideoZoomer~\cite{ding2025videozoomer} use zoom-in or temporal
focusing; VideoChat-R1.5~\cite{videochatr152025} performs iterative
perception; Video-o3~\cite{zeng2026videoo3},
Open-o3 Video~\cite{openo3video2025}, and
VideoTemp-o3~\cite{videotempo32026} explore native interleaved tool
invocation. These methods show that active evidence acquisition is
important for long-video reasoning. However, their retrieval granularity is still largely governed by
external tools, preset zoom/spotlight modes, or system-defined
visual-token budgets. As a result, obtaining the right amount of visual
evidence for each question often depends on repeated retrieval calls,
which expand the inference context and make training harder.

\paragraph{Frame sampling for video MLLMs.}
Frame sampling determines which visual evidence is available to the
reasoner and directly affects both accuracy and efficiency. Uniform
sampling at a fixed interval~\cite{maaz2023video,zhang2023videollama,
jin2024chat} is simple but content-agnostic, so it can miss short
events in long videos. Modern video MLLMs~\cite{bai2025qwen2.5vl,
bai2025qwen3vl} provide dynamic-FPS or timestamp-aligned video
interfaces, but the sampling rate is set by the calling pipeline
rather than predicted by the model during reasoning. Query-conditioned
selectors such as AKS~\cite{tang2025aks}, FOCUS~\cite{yao2025focus},
and Frame-Voyager~\cite{yu2024framevoyager} improve over uniform
sampling by ranking frames according to query relevance, but they
usually commit to a fixed frame set before reasoning begins. Agentic
video systems lift this one-shot constraint by allowing inference-time
retrieval, yet their visual budget is still largely controlled by
system-level schedules, fixed per-call caps, slow/fast presets, or
external visual-token quotas.

\paragraph{Reinforcement learning for MLLM reasoning.}
GRPO from DeepSeek-R1~\cite{guo2025deepseekr1,
shao2024deepseekmath} has been adopted to post-train MLLMs for
image VQA~\cite{zhang2025r1vl,huang2025visionr1}, video
reasoning~\cite{feng2025videor1,grpocare2025,videorts2025},
and tool-augmented
generation~\cite{zheng2025deepeyes,vital2025}. These formulations
apply a single trajectory-level advantage to every token in
the rollout, entangling the credit for committing a retrieval
action with the credit for producing the final answer.
Variants that decouple along task
difficulty~\cite{vital2025}, multi-step vs.\ single-step
turn~\cite{lover12025}, or reward
component~\cite{yang2025longvt} still share one advantage
across all tokens within a rollout, leaving the
\emph{where-to-look} decision without a dedicated training
signal.

\begin{figure*}[!t]
    \centering
    \includegraphics[width=\textwidth]{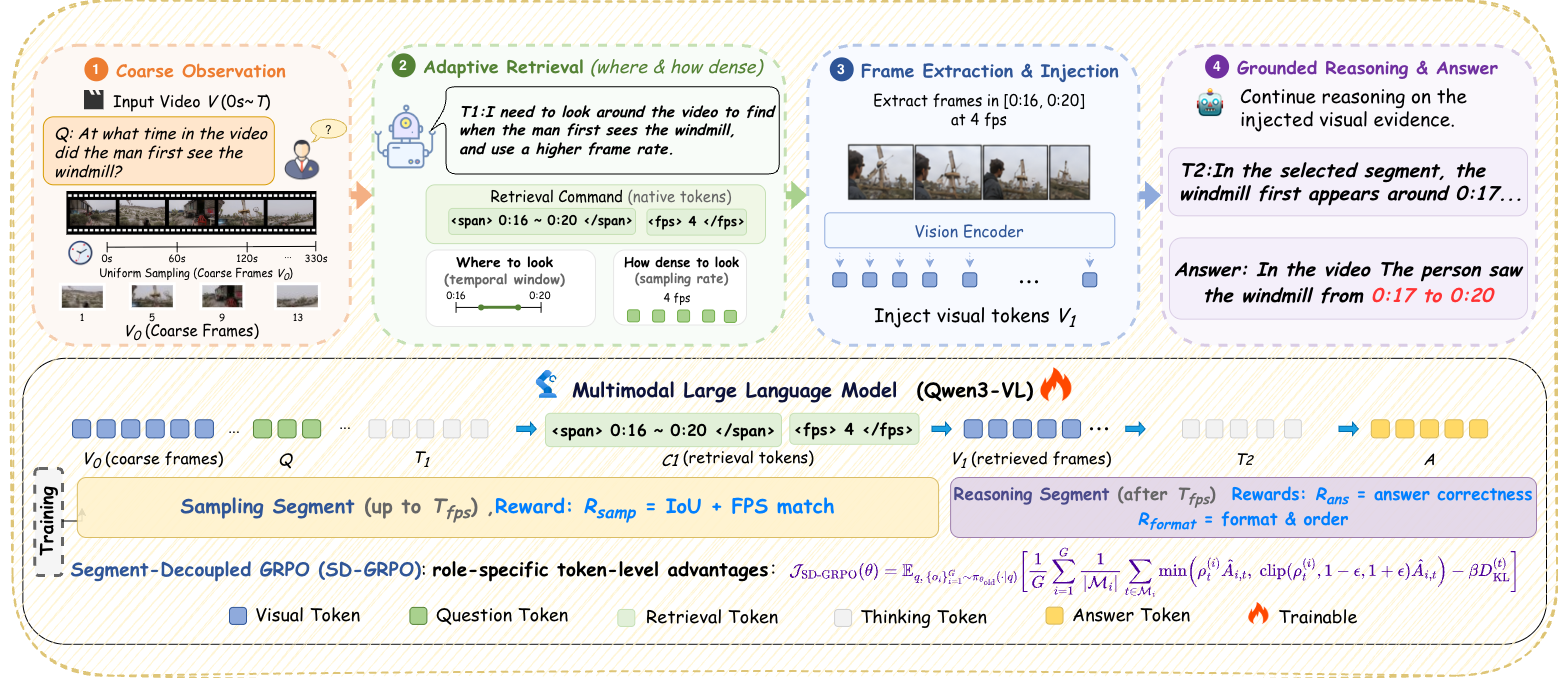}
    \caption{\textbf{Overview of DynFrame.}
    The model interleaves tokenized temporal retrieval
    (\texttt{<span>}, \texttt{<fps>}) with on-the-fly frame
    injection inside a single autoregressive pass. SD-GRPO splits
    each rollout at the retrieval boundary and applies
    segment-specific advantages so that the sampling decision and
    the answer reasoning are credited separately.}
    \label{fig:framework}
\end{figure*}

%% file: sections/03_method.tex
\section{Method}
\label{sec:method}

We propose \textbf{DynFrame}, an end-to-end trainable video
reasoning framework that unifies adaptive frame retrieval with
dynamically interleaved vision-language reasoning within a single
generative process (Fig.~\ref{fig:framework}). Built upon
Qwen3-VL~\cite{bai2025qwen3vl}, DynFrame introduces three key
designs: (i) a \emph{tokenized retrieval interface}, where the
model specifies which temporal span and at what sampling density
to retrieve by generating structured tokens; (ii) a \emph{dynamic
frame injection} mechanism that encodes the retrieved frames and
inserts them into the decoding context on-the-fly; and (iii)
\emph{Segment-Decoupled GRPO}, which decouples rewards across
response segments to separately optimize temporal selection and
answer reasoning. We detail the dynamic multimodal CoT in
\S\ref{sec:dmcot}, the two-stage training procedure in
\S\ref{sec:training}, and the dataset curation pipeline in
\S\ref{sec:data}.

\subsection{Dynamic Multimodal Chain-of-Thought}
\label{sec:dmcot}

\subsubsection{Generation with Adaptive Frame Retrieval}
\label{sec:gen_retr}

Given a user question $Q$ and an initial video observation $V_0$
(uniformly sampled frames), DynFrame generates a multimodal reasoning
trajectory that interleaves textual reasoning with adaptive retrieval.
In our main setting, this forms a three-stage process:
\emph{coarse reasoning} $\rightarrow$ \emph{retrieval}
$\rightarrow$ \emph{grounded reasoning}. The model first produces an
initial reasoning segment $T_1$ based on $Q$ and a coarse understanding
of $V_0$, then generates a retrieval command $C_1$ to request
additional evidence from the original video. After the retrieved frames
$V_1$ are injected, the model continues reasoning over the augmented
visual context and produces the final answer $A$. The trajectory is:
\begin{equation}
\mathbf{s} = \{V_0,\, T_1,\, C_1,\, V_1,\, T_2,\, A\}.
\end{equation}
In this work, we instantiate a single retrieval round, which provides a
favorable accuracy--cost trade-off on our benchmarks. The same
tokenized interface can be extended to multiple rounds by emitting
additional retrieval commands.

\noindent\textbf{Tokenized retrieval interface.}
We design a set of special tokens to express video evidence acquisition
as part of the model output. The model emits
\texttt{<span>}$t_s$--$t_e$\texttt{</span>} to specify a temporal
window and \texttt{<fps>}$f$\texttt{</fps>} to specify the sampling
frame rate, which together parameterize the retrieval command $C_1$.
This turns frame selection from a fixed preprocessing choice into a
learnable decision within the autoregressive generation trajectory,
enabling adaptive, multi-granularity temporal sampling conditioned on
the current reasoning context. Predicting both the temporal window and
the sampling rate is essential, as different queries demand different
granularities: a brief hand gesture requires dense frames within a
narrow window, whereas long-form narrative understanding may only need
sparse keyframes.

\subsubsection{Dynamic Frame Injection}
\label{sec:inject}

We perform on-the-fly frame injection during autoregressive generation.
The \texttt{</fps>} token closing a retrieval command acts as a
retrieval trigger: the system parses the preceding \texttt{<span>} and
\texttt{<fps>} fields to obtain $(t_s,t_e,f)$ and extracts
\begin{equation}
N = \left\lfloor (t_e - t_s)\times f \right\rfloor
\end{equation}
frames uniformly distributed within $[t_s,t_e]$, capped by a maximum
frame budget via uniform subsampling when necessary, and independent of
the initial uniform sampling. Retrieved frames pass through the frozen
vision encoder, and for each frame we emit a timestamped visual token
subsequence containing its timestamp, vision boundary tokens, and
$H'W'/m^2$ visual placeholder tokens, where $H',W'$ are the
post-encoder spatial dimensions and $m$ is the patch merge size. The
new visual tokens are appended to the generation buffer, yielding an
extended sequence
$\mathbf{x}'=[\mathbf{x}_{1:L};\,\mathbf{v}_{1:N}]$. The decoder then
re-prefills KV states over $\mathbf{x}'$ before incremental decoding
resumes:
\begin{equation}
\mathbf{H}_l = \mathrm{SelfAttn}_l(\mathbf{x}'),
\quad l=1,\ldots,L_{\text{dec}},
\end{equation}
so that the inserted frames can attend to all prior reasoning tokens
and vice versa. This bidirectional attention lets the post-injection
reasoning ground its claims directly on the retrieved frames rather
than on a textual restatement of them.

\subsection{Training Framework}
\label{sec:training}

Our training procedure consists of two stages: (1) cold-start
supervised fine-tuning (SFT) to establish interleaved retrieval and
reasoning behaviors, and (2) reinforcement learning via SD-GRPO to
jointly improve temporal selection and grounded multimodal reasoning.
We train on two curated multi-task datasets. \textbf{DM-CoT-74k}
provides supervised trajectories with interleaved retrieval commands
(span/FPS) and grounded reasoning. \textbf{DM-RL-45k} provides
question--video pairs with ground-truth temporal spans, FPS targets,
and answers, enabling reward computation for both sampling quality and
answer correctness. Full dataset details are in \S\ref{sec:data}.

\subsubsection{Cold-Start Supervised Fine-Tuning}
\label{sec:sft}

We bootstrap DynFrame with SFT on \textbf{DM-CoT-74k} by maximizing
the likelihood of interleaved reasoning and retrieval tokens:
\begin{equation}
\mathcal{L}_{\text{SFT}} =
-\frac{1}{|\mathcal{M}|} \sum_{t\in\mathcal{M}}
\log p_\theta(x_t \mid x_{<t}),
\end{equation}
where $\mathcal{M}=\{t: x_t \neq \texttt{<|video\_pad|>}\}$ denotes
positions whose targets are not visual placeholder tokens. We exclude
\texttt{<|video\_pad|>} tokens within injected segments from the loss,
as they correspond to vision features rather than predicted text. In
contrast, we retain timestamp tokens and vision boundary tokens
(e.g., \texttt{<|vision\_start|>}, \texttt{<|vision\_end|>}) as SFT
targets, which encourages the model to learn temporal boundary
prediction.

\subsubsection{Segment-Decoupled GRPO}
\label{sec:sdgrpo}

While SFT teaches the model to imitate interleaved
retrieval--reasoning trajectories, we observe a
\emph{credit-assignment imbalance} when directly applying GRPO to our
retrieval-augmented generation. On challenging questions, the model
often predicts a reasonable span/FPS but still fails in post-injection
reasoning, causing the negative outcome reward to penalize the
retrieval tokens as well. Conversely, when the final answer can be
obtained from coarse initial observations (e.g., shortcut cues), a
positive outcome reward may incorrectly reinforce inaccurate span/FPS
predictions. To address this, we propose
\textbf{Segment-Decoupled GRPO (SD-GRPO)}, which separates
optimization for the \emph{sampling segment} (span/FPS tokens before
injection) from the \emph{grounded reasoning segment} (tokens after
injection). SD-GRPO assigns a retrieval-specific reward to the sampling
segment and an answer-specific reward to the post-injection segment,
improving credit assignment and stabilizing retrieval--reasoning
co-adaptation.

We extend standard GRPO~\cite{shao2024deepseekmath} rollouts to
accommodate dynamic video insertion. Given a question $Q$ and video
$V$, we sample $G$ completions $\{o_1,\ldots,o_G\}$ from the current
policy $\pi_\theta$. When the model generates the retrieval terminator
token \texttt{</fps>} during a rollout, the system triggers dynamic
frame injection and appends the corresponding visual token sequences
to the generation buffer. We denote the position of \texttt{</fps>} as
$T_{\text{fps}}$, which partitions each completion into a
\textit{sampling segment}
$o^{\text{samp}}=\{o_t\}_{t=1}^{T_{\text{fps}}}$ and a
\textit{reasoning segment}
$o^{\text{reas}}=\{o_t\}_{t=T_{\text{fps}}+1}^{T}$. We define three
reward signals.

\noindent\textbf{(1) Sampling reward.}
The sampling reward $R_{\text{samp}}$ evaluates the quality of the
frame acquisition decision, combining temporal span overlap with a
smooth FPS matching score:
\begin{equation}
\label{eq:samp_reward}
R_{\text{samp}}
= \lambda_1 \cdot \text{IoU}\big([\hat{t}_s, \hat{t}_e],\;
[t_s^*, t_e^*]\big)
+ \lambda_2 \cdot \max\!\left(0,\;
1 - \frac{|\hat{f} - f^*|}{f_{\max}}\right),
\end{equation}
where $[\hat{t}_s, \hat{t}_e]$ and $\hat{f}$ are the model's
predictions, $[t_s^*, t_e^*]$ and $f^*$ are the ground-truth
annotations, and the FPS term decays linearly from full credit at
$\hat{f}=f^*$ to zero at deviations exceeding $f_{\max}$. Both terms
are bounded in $[0,1]$, so we combine them at equal scale with
$\lambda_1{=}\lambda_2{=}0.5$. The two terms play asymmetric roles in
practice: span IoU localizes \emph{where} evidence lies and is the
dominant supervisor, while the FPS term acts as a fine-grained
corrector that selects density \emph{within} an already-localized
window. This is consistent with the intuition that any FPS choice is
low-utility once the span is wrong, and is verified empirically in
\S\ref{sec:ablation}.

\noindent\textbf{(2) Answer reward.}
The answer reward $R_{\text{ans}}$ evaluates final-task correctness: we use
exact match for multiple-choice VideoQA, and IoU for
temporal grounding.

\noindent\textbf{(3) Format reward.}
We additionally apply a rule-based format reward $R_{\text{format}}$ to ensure
the output follows the required structure (reasoning segment, span/FPS fields,
and answer segment with correct ordering and pairing).

\noindent\textbf{Token-Level Segment-Decoupled Advantage.}
The key idea of SD-GRPO is to assign advantages according to which segment a
token belongs to, rather than using a single scalar advantage for the whole
completion. For each group of $G$ rollouts, we compute two group-normalized
advantages:
\begin{equation}
    \hat{A}_i^{\text{samp}} =
    \frac{R_{\text{samp},i} - \mu_{\text{samp}}}{\sigma_{\text{samp}} + \epsilon},
    \qquad
    \hat{A}_i^{\text{ans}} =
    \frac{(R_{\text{ans},i} + R_{\text{format},i}) - \mu_{\text{ans}}}{\sigma_{\text{ans}} + \epsilon},
\end{equation}
where $\mu$ and $\sigma$ are computed within the group for each reward. We then
assign per-token advantages by:
\begin{equation}
    \hat{A}_{i,t} =
    \begin{cases}
        \hat{A}_i^{\text{samp}} + \hat{A}_i^{\text{ans}}
        & \text{if } t \leq T_{\text{fps}} \quad \text{(sampling segment)} \\[4pt]
        \hat{A}_i^{\text{ans}}
        & \text{if } t > T_{\text{fps}} \quad \text{(reasoning segment)}.
    \end{cases}
\end{equation}
Intuitively, tokens before \texttt{</fps>} are directly responsible for span/FPS
decisions and thus receive $R_{\text{samp}}$-based credit, while also receiving
an end-to-end signal via $R_{\text{ans}}$. Tokens after \texttt{</fps>} cannot
change the already-committed sampling decision, and are optimized solely for
answer correctness.

\noindent\textbf{Optimization.}
With token-level segment-decoupled advantages, we optimize the SD-GRPO objective as follows:
\begin{align}
\mathcal{J}_{\text{SD-GRPO}}(\theta) &=
\mathbb{E}_{q,\, \{o_i\}_{i=1}^G \sim \pi_{\theta_{\text{old}}}(\cdot|q)} \notag \\
\bigg[
\frac{1}{G} \sum_{i=1}^G \frac{1}{|\mathcal{M}_i|} &\sum_{t \in \mathcal{M}_i}
\min\!\big(\rho_t^{(i)} \hat{A}_{i,t},\;
\text{clip}(\rho_t^{(i)}, 1{-}\epsilon, 1{+}\epsilon)\hat{A}_{i,t}\big)
- \beta\, D_{\text{KL}}^{(t)}
\bigg],
\end{align}
where $q=\{Q,V\}$ denotes the input question and video, and $\{o_i\}_{i=1}^G$ are
$G$ rollouts sampled from the behavior policy $\pi_{\theta_{\text{old}}}$.
$\mathcal{M}_i=\{t : o_{i,t} \notin \mathcal{V}_{\text{pad}}\}$ excludes visual
placeholder tokens (e.g., \texttt{<|video\_pad|>}) from optimization. The
importance ratio is $\rho_t^{(i)}=\pi_{\theta}(o_{i,t}\mid q,o_{i,<t}) /
\pi_{\theta_{\text{old}}}(o_{i,t}\mid q,o_{i,<t})$, and $D_{\text{KL}}^{(t)}$ is
the per-token KL divergence against the reference policy $\pi_{\text{ref}}$ with
coefficient $\beta$. Compared to the standard GRPO that uses a single trajectory-level
advantage for all tokens, SD-GRPO assigns segment-dependent token-level
advantages $\hat{A}_{i,t}$, enabling targeted optimization for both
\textit{where to look} (sampling segment) and \textit{how to reason} (reasoning
segment).

\subsection{Training Data Curation}
\label{sec:data}

\begin{figure*}[!t]
    \centering
    \includegraphics[width=0.95\textwidth]{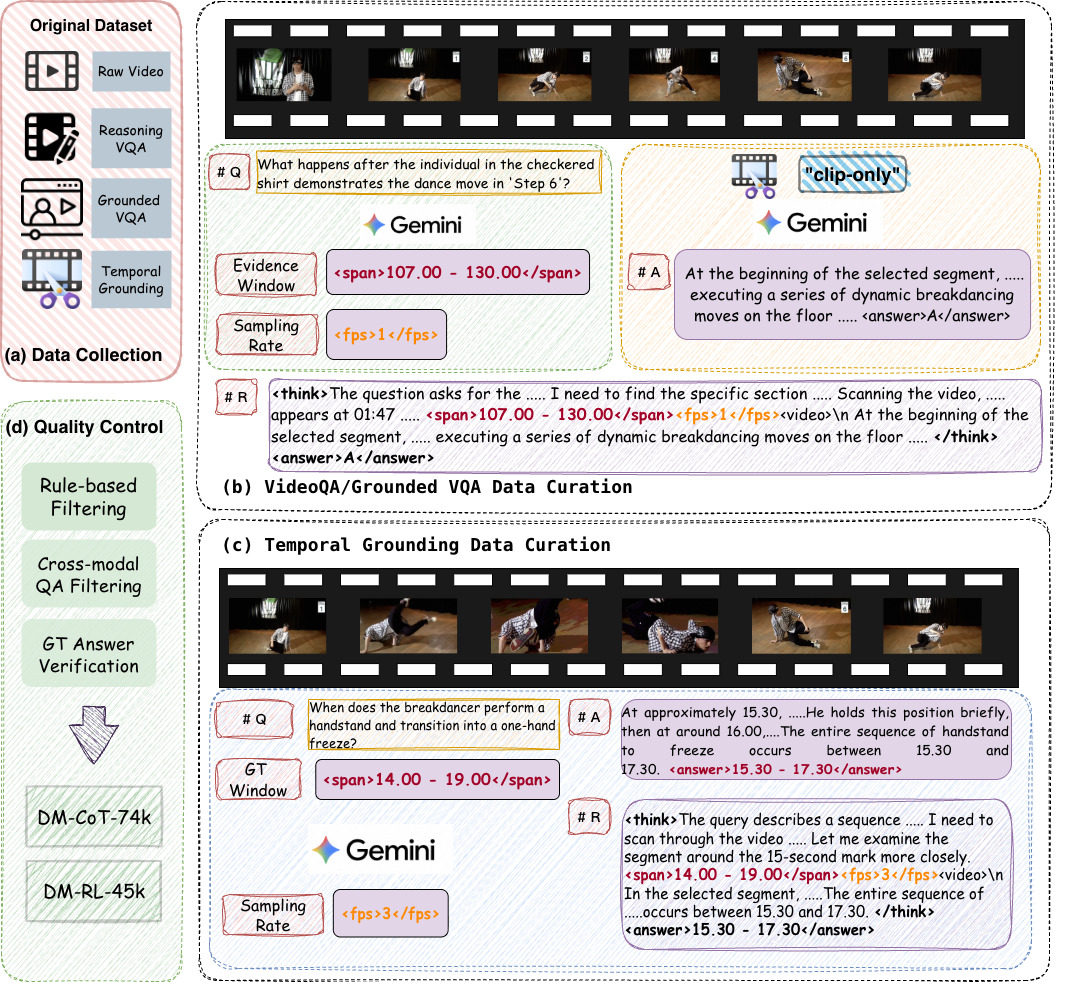}
    \caption{\textbf{Data curation pipeline for DM-CoT-74k and DM-RL-45k.}
    (a)~Sources: VideoQA, grounded VideoQA, and temporal grounding benchmarks.
    (b)~For VideoQA without temporal annotations, Gemini selects the evidence
    window and sampling rate, then answers under a ``clip-only'' constraint
    enforced at the prompt level.
    (c)~For temporal grounding, ground-truth windows are reused; Gemini only
    selects an activity-adaptive FPS.
    (d)~Rule-based and answer-consistency filters yield the final mixtures.}
    \label{fig:datapipeline}
\end{figure*}

Training DynFrame requires trajectories that interleave textual
reasoning with span+FPS retrieval commands and grounded answers.
Since no public dataset provides this format, we curate two
task-balanced mixtures over temporal grounding, VideoQA, and
grounded VideoQA, sourced from
Charades-STA~\cite{song2024temporalgrounding},
ActivityNet-MR~\cite{krishna2017dense},
Video-R1~\cite{feng2025videor1},
ReXTime~\cite{chen2024rextime}, and NExT-GQA~\cite{nextgqa}. All
samples are converted into the unified retrieval-augmented format
of \S\ref{sec:gen_retr}.

For VideoQA and grounded VideoQA, where temporal annotations are
absent, we prompt
Gemini-3-Pro~\cite{gemini3pro2026} to identify the
relevant evidence window, propose an FPS, and produce the answer
under the constraint that it ``only rewatched the proposed
window''---enforced at the prompt level rather than by
re-uploading trimmed clips, which keeps construction cost low.
For temporal grounding, the human-annotated boundary is reused
verbatim with a $0.5$--$2$\,s random margin to provide context,
and Gemini is queried only for an activity-adaptive FPS conditioned
on the span. Because no public video dataset provides per-segment
FPS supervision, the FPS targets used during RL are inherited from
this teacher; the temporal IoU term in
$R_{\text{samp}}$ remains anchored to human-annotated boundaries,
which is consistent with our design of treating IoU as the
dominant supervisory signal.

We then apply two filtering stages: rule-based checks discard
samples with missing or reversed retrieval fields, and an
answer-consistency check drops teacher answers that disagree with
the ground-truth label. Together they remove roughly $40\%$ of
raw teacher outputs. The final 74k SFT mixture comprises
${\sim}30\%$ temporal grounding, ${\sim}45\%$ VideoQA, and
${\sim}25\%$ grounded VideoQA; DM-RL-45k follows a similar ratio.
Detailed prompts used for data generation are provided in 
Appendix~\ref{sec:appendix_prompts}.

%% file: sections/04_experiments.tex
\section{Experiments}
\label{sec:exp}

\subsection{Experimental Setup}
\label{sec:setup}

\noindent\textbf{Benchmarks.}
We evaluate on six benchmarks spanning three task families:
(i)~\emph{grounded VideoQA}---NExT-GQA~\cite{nextgqa}, with
answer accuracy (Acc) and grounding mIoU;
(ii)~\emph{temporal sentence
grounding}---Charades-STA~\cite{gao2017tall} and
ActivityNet-MR~\cite{krishna2017dense}, with
R@$\{0.3,0.5,0.7\}$ and mIoU;
(iii)~\emph{long-form video understanding}---Video-MME (w/o
sub.)~\cite{fu2025videomme}, MLVU
(M-Avg)~\cite{mlvu}, and LVBench~\cite{wang2024lvbench}, all
measured by multi-choice accuracy.

\noindent\textbf{Implementation.}
We build DynFrame on Qwen3-VL-Thinking at 4B and 8B with the
visual encoder frozen. The initial pass uniformly samples at
$f_1{=}2$\,fps, and the adaptive retrieval round uses a
model-predicted $f_2\!\in\![1,6]$\,fps, with up to $N{=}128$
frames retrieved. SFT runs $4{,}000$ steps at lr
$1\!\times\!10^{-5}$, batch $256$, on $64$ H200 GPUs (AdamW); RL
with SD-GRPO uses lr $1\!\times\!10^{-6}$, group $G{=}8$,
temperature $1.0$, for $1{,}000$ further steps. Detailed
inference-cost comparisons across benchmarks and methods are
provided in Appendix~\ref{app:cost_protocol}.

\begin{table*}[!htb]
\centering
\renewcommand{\arraystretch}{1.12}
\caption{\textbf{Comparison of our method with existing methods across six benchmarks.}
``--'' indicates the original paper does not evaluate on that benchmark, or the
model's output format is incompatible with the metric.}
\label{tab:main_combined}
\resizebox{\textwidth}{!}{%
\begin{tabular}{l|c|cc|cccc|cccc|ccc}
\toprule[1pt]
\multirow{2}{*}{\textbf{Model}}
 & \multirow{2}{*}{\textbf{Size}}
 & \multicolumn{2}{c|}{\textbf{NExT-GQA}}
 & \multicolumn{4}{c|}{\textbf{Charades-STA}}
 & \multicolumn{4}{c|}{\textbf{ActivityNet-MR}}
 & \textbf{V-MME} & \textbf{MLVU} & \textbf{LVB} \\
\cmidrule(lr){3-4}\cmidrule(lr){5-8}\cmidrule(lr){9-12}
 & & Acc & mIoU
 & R@.3 & R@.5 & R@.7 & mIoU
 & R@.3 & R@.5 & R@.7 & mIoU
 & w/o sub & M-Avg & Acc \\
\midrule\midrule

\multicolumn{15}{c}{\textbf{\textit{General / Single-turn Video MLLMs}}} \\
\midrule
Qwen2.5-VL~\cite{bai2025qwen2.5vl}
 & 7B
 & 76.5 & 30.5
 & 64.7 & 43.1 & 22.8 & 43.6
 & 41.6 & 23.2 & 9.5 & 28.9
 & 65.1 & 70.2 & 45.3 \\

InternVL3~\cite{zhu2025internvl3}
 & 8B
 & \underline{\textbf{80.4}} & 30.0
 & -- & -- & -- & --
 & -- & -- & -- & --
 & 66.3 & 71.4 & 47.0 \\

Qwen3-VL (Thinking)~\cite{bai2025qwen3vl}
 & 4B
 & 73.8 & 33.4
 & 80.5 & 69.2 & 44.6 & 59.0
 & 49.8 & 30.4 & 14.1 & 34.5
 & 68.9 & 75.7 & 53.5 \\

Qwen3-VL (Thinking)~\cite{bai2025qwen3vl}
 & 8B
 & 75.4 & 35.1
 & 81.6 & 70.8 & 45.8 & 59.9
 & 51.8 & 32.4 & 15.6 & 36.2
 & 71.8 & 75.1 & 55.8 \\

Video-R1-7B~\cite{feng2025videor1}
 & 7B
 & 77.3 & --
 & 63.5 & 44.0 & 22.5 & 43.8
 & 40.0 & 22.5 & 9.5 & 28.5
 & 61.4 & -- & -- \\

\midrule\midrule
\multicolumn{15}{c}{\textbf{\textit{Thinking-with-Video / Tool-Augmented Methods}}} \\
\midrule
Temporal-RLT~\cite{li2025temporalrlt}
 & 7B
 & 78.7 & 37.3
 & 79.6 & 67.9 & 44.1 & 57.0
 & 56.9 & 38.4 & 20.2 & 39.0
 & 57.6 & -- & -- \\

VITAL~\cite{vital2025}
 & 7B
 & 78.7 & 43.0
 & 83.1 & 72.0 & 46.7 & 59.9
 & 70.9 & 50.8 & 31.6 & 49.8
 & 64.1 & -- & -- \\

LongVT~\cite{yang2025longvt}
 & 7B
 & 70.4 & 17.4
 & 41.0 & 25.8 & 11.7 & 27.2
 & 32.4 & 18.6 & 9.2 & 20.5
 & -- & -- & 41.3 \\

VideoZoomer~\cite{ding2025videozoomer}
 & 7B
 & -- & --
 & -- & -- & -- & --
 & -- & -- & -- & --
 & 65.2 & 68.8 & 41.5 \\

LOVE-R1~\cite{lover12025}
 & 7B
 & 73.0 & 30.5
 & 74.0 & 41.0 & 14.0 & 44.8
 & 49.0 & 24.0 & 13.0 & 30.4
 & 66.2 & 67.4 & 48.2 \\

VideoChat-R1.5~\cite{videochatr152025}
 & 7B
 & 79.9 & --
 & 82.8 & 71.6 & 48.3 & 60.6
 & 52.4 & 32.3 & 16.8 & 35.5
 & 67.1 & 70.9 & 48.4 \\

Video-o3~\cite{zeng2026videoo3}
 & 7B
 & -- & --
 & 83.3 & 71.9 & 49.0 & 60.7
 & -- & -- & -- & --
 & 66.5 & 72.1 & 47.6 \\

\midrule\midrule
\rowcolor{gray!18}
\textbf{DynFrame-4B (ours)}
 & 4B
 & 77.6 & 41.5
 & 83.5 & 71.0 & 46.5 & 60.0
 & 70.4 & 49.2 & 28.6 & 47.5
 & 69.5 & 76.3 & 54.8 \\

\rowcolor{gray!18}
\textbf{DynFrame-8B (ours)}
 & 8B
 & 80.0 & \underline{\textbf{44.3}}
 & \underline{\textbf{85.1}} & \underline{\textbf{72.5}}
 & \underline{\textbf{49.4}} & \underline{\textbf{61.7}}
 & \underline{\textbf{72.4}} & \underline{\textbf{52.0}}
 & \underline{\textbf{33.1}} & \underline{\textbf{51.5}}
 & \underline{\textbf{72.3}} & \underline{\textbf{77.1}} & \underline{\textbf{56.9}} \\
\bottomrule[1pt]
\end{tabular}
}
\vspace{2pt}

\end{table*}

\subsection{Comparison with State-of-the-Art}
\label{sec:main_results}

\noindent\textbf{Temporal sentence grounding.}
On Charades-STA, DynFrame-8B reaches a new state of the art at
$61.7$\,mIoU, surpassing the previous best thinking-with-video
method Video-o3 ($60.7$); the 4B variant remains competitive at
$60.0$\,mIoU, matching VITAL-7B ($59.9$) at half the parameter
count. On the more challenging ActivityNet-MR, DynFrame-8B
improves over the strongest baseline VITAL-7B by $+1.7$\,mIoU.
These gains show that a single round of joint span--density
retrieval can match or surpass multi-round tool-call methods,
with larger advantages on longer videos.

\noindent\textbf{Grounded VideoQA.}
On NExT-GQA, DynFrame-8B achieves the best joint score among
grounding-capable models ($80.0$\,Acc / $44.3$\,mIoU),
improving over VITAL-7B by $+1.3$ on both metrics and over its
Qwen3-VL-Thinking-8B backbone by $+4.6$\,Acc / $+9.2$\,mIoU.
Although InternVL3-8B reports a slightly higher accuracy
($80.4$), its $30.0$\,mIoU shows that the correct answers are
not visually grounded. The simultaneous improvement on both
metrics confirms that SD-GRPO effectively credits the sampling
decision and the answer-reasoning segment separately.

\noindent\textbf{Long-form video understanding.}
On Video-MME / MLVU / LVBench, DynFrame-8B sets new best results
across all three long-form benchmarks ($72.3$\,/\,$77.1$\,/\,$56.9$),
improving over its strong Qwen3-VL-Thinking-8B backbone by
$+0.5$\,/\,$+2.0$\,/\,$+1.1$. The 4B variant also surpasses every 7B
tool-augmented method (LOVE-R1, VideoChat-R1.5, Video-o3) on
long-form video. Together with the grounding and grounded-VideoQA
results, these findings show that learnable span--density retrieval
brings complementary gains across short-form grounding, grounded
VideoQA, and long-form video understanding.

\subsection{Ablation Study}
\label{sec:ablation}

\begin{table}[!htb]
\centering
\caption{\textbf{Ablation studies on the 8B model.}
NExT-GQA and Charades-STA report mIoU; V-MME and LVBench report accuracy.
Shaded rows are our default. (a)~Masking strategy for injected frames.
(b)~Effectiveness of SD-GRPO. (c)~Robustness to the initial sampling
rate $f_1$. (d)~Effectiveness of dynamic retrieval FPS $f_2$.}
\label{tab:ablation}
\footnotesize
\setlength{\tabcolsep}{2.6pt}
\renewcommand{\arraystretch}{1.05}

\begin{subtable}[t]{0.49\linewidth}
\centering
\caption{Masking strategy.}
\label{tab:abl_mask}
\begin{tabular}{lcccc}
\toprule
Variant & N-GQA & Cha & V-MME & LVB \\
\midrule
Mask all video tokens & 18.1 & 26.4 & 36.3 & 39.4 \\
\rowcolor{gray!18}
Mask video pad only & \textbf{44.3} & \textbf{61.7} & \textbf{72.3} & \textbf{56.9} \\
\bottomrule
\end{tabular}
\end{subtable}\hfill%
\begin{subtable}[t]{0.49\linewidth}
\centering
\caption{Effectiveness of SD-GRPO.}
\label{tab:abl_sdgrpo}
\begin{tabular}{lcccc}
\toprule
Variant & N-GQA & Cha & V-MME & LVB \\
\midrule
SFT only        & 41.5 & 58.7 & 70.0 & 54.5 \\
+ vanilla GRPO  & 42.2 & 59.5 & 70.3 & 54.8 \\
\rowcolor{gray!18}
+ SD-GRPO       & \textbf{44.3} & \textbf{61.7} & \textbf{72.3} & \textbf{56.9} \\
\bottomrule
\end{tabular}
\end{subtable}

\vspace{2pt}

\begin{subtable}[t]{0.49\linewidth}
\centering
\caption{Initial sampling rate $f_1$.}
\label{tab:abl_f1}
\begin{tabular}{lcccc}
\toprule
Variant & N-GQA & Cha & V-MME & LVB \\
\midrule
Qwen3-VL, $f_1{=}0.5$ & 23.7 & 46.3 & 58.2 & 45.0 \\
Qwen3-VL, $f_1{=}2$   & 35.1 & 59.9 & 71.8 & 55.8 \\
DynFrame, $f_1{=}0.5$, $f_2{=}$dyn & 42.5 & 60.5 & 70.5 & 55.1 \\
\rowcolor{gray!18}
DynFrame, $f_1{=}2$, $f_2{=}$dyn   & \textbf{44.3} & \textbf{61.7} & \textbf{72.3} & \textbf{56.9} \\
\bottomrule
\end{tabular}
\end{subtable}\hfill%
\begin{subtable}[t]{0.49\linewidth}
\centering
\caption{Retrieval FPS $f_2$.}
\label{tab:abl_f2}
\begin{tabular}{lcccc}
\toprule
Variant & N-GQA & Cha & V-MME & LVB \\
\midrule
DynFrame, $f_2{=}2$ fixed & 41.9 & 58.9 & 70.2 & 55.8 \\
\rowcolor{gray!18}
DynFrame, $f_2{=}$dyn     & \textbf{44.3} & \textbf{61.7} & \textbf{72.3} & \textbf{56.9} \\
\bottomrule
\end{tabular}
\end{subtable}
\end{table}

\begin{figure}[!htb]
\centering
\begin{subfigure}[t]{0.44\linewidth}
\centering
\includegraphics[width=\linewidth]{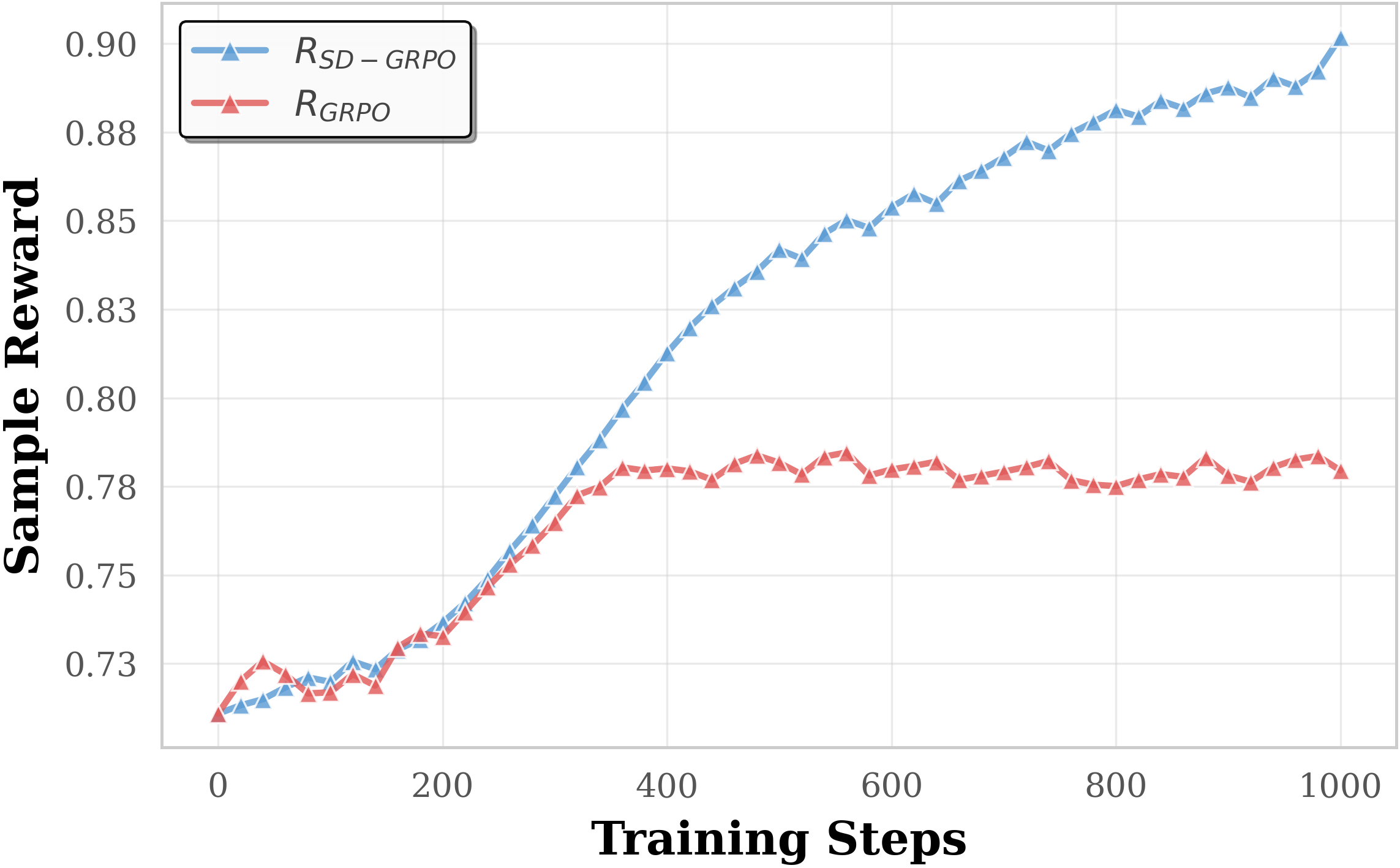}
\caption{$R_{\text{samp}}$ vs.\ steps.}
\label{fig:reward_curve_a}
\end{subfigure}\hspace{8pt}%
\begin{subfigure}[t]{0.44\linewidth}
\centering
\includegraphics[width=\linewidth]{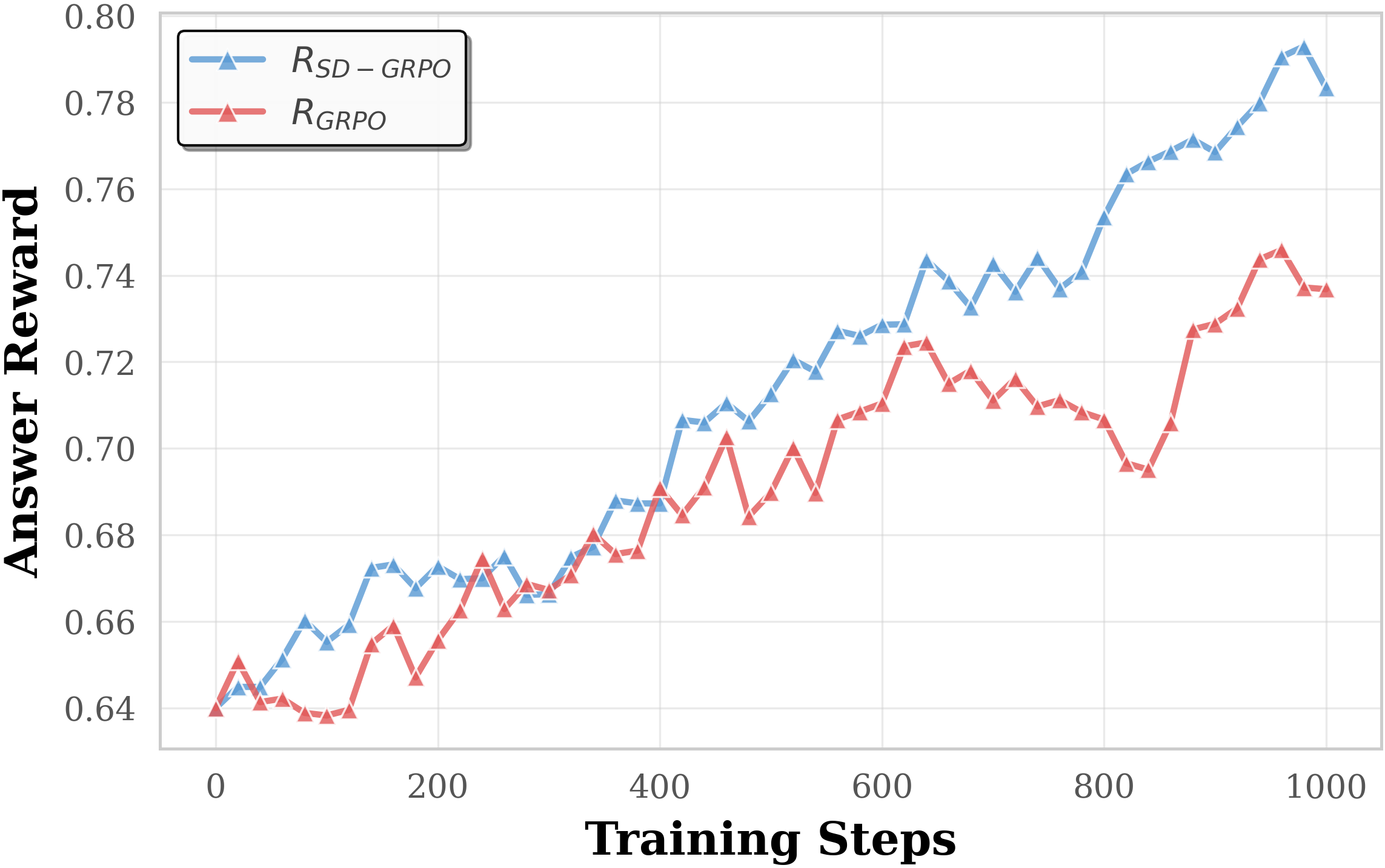}
\caption{$R_{\text{ans}}$ vs.\ steps.}
\label{fig:reward_curve_b}
\end{subfigure}
\caption{\textbf{Reward dynamics during RL.} SD-GRPO lifts both
the sampling reward $R_{\text{samp}}$ and the answer reward
$R_{\text{ans}}$ over vanilla GRPO.}
\label{fig:reward_curve}
\end{figure}

\noindent\textbf{(a) Masking strategy.}
During SFT, only the visual placeholder tokens
(\texttt{<|video\_pad|>}) are excluded from the loss because they
encode raw vision features rather than predicted text, while
timestamps and vision-boundary markers
(\texttt{<|vision\_start|>}, \texttt{<|vision\_end|>}) are kept as
supervision targets. Masking these additional tokens causes a sharp
drop in performance, confirming that they act as anchors that align
the injected frames with the reasoning context.
\textbf{(b) SD-GRPO.}
Building on this SFT initialization, we next examine the effect
of segment-decoupled RL. SFT $\rightarrow$ vanilla GRPO
$\rightarrow$ SD-GRPO improves metrics monotonically (Charades
mIoU $58.7\!\to\!59.5\!\to\!61.7$), with SD-GRPO consistently
outperforming vanilla GRPO across all four benchmarks.
Fig.~\ref{fig:reward_curve} further shows that SD-GRPO yields
higher sampling and answer rewards throughout training,
confirming that segment-level credit assignment effectively
decouples the sampling decision from the answer reasoning.
\textbf{(c) Robustness to $f_1$.}
Beyond training-time choices, we also study how the model behaves
under tighter initial frame budgets. Reducing $f_1$ from 2 to 0.5 fps causes the Qwen3-VL-Thinking-8B
backbone to degrade substantially, with drops of up to 13.6 points
across the four benchmarks. In contrast, DynFrame degrades by at most
1.8 points, suggesting that dynamic retrieval can recover most of the
evidence missed by a sparse initial pass.
\textbf{(d) Dynamic FPS.}
Finally, we verify the necessity of letting the model choose its
retrieval frame rate. Replacing the model-predicted $f_2$ with a
fixed $2$\,fps rate consistently reduces performance by
$1.1$\,--\,$2.8$ points across all four benchmarks, confirming
that DynFrame can adaptively select different sampling densities
for different questions, acquiring evidence at the appropriate
granularity. Detailed analysis of predicted spans and FPS is
provided in Appendix~\ref{app:span_fps_analysis}.

%% file: sections/07_conclusion.tex
\enlargethispage{3\baselineskip}
\vspace{-0.6em}
\section{Conclusion}
\label{sec:conclusion}
\vspace{-0.3em}

We presented DynFrame, a dynamic multimodal reasoning
framework that turns visual evidence acquisition into a model-native
decision. By predicting both the temporal window and the sampling
density, DynFrame jointly decides where to retrieve and
how densely to sample, acquiring task-adaptive,
multi-granularity evidence with a single retrieval step. To train this
behavior, we curated DM-CoT-74k and DM-RL-45k, and
introduced Segment-Decoupled GRPO, which separately credits
the sampling decision and the answer reasoning. Experiments across
grounded VideoQA, temporal grounding, and long-form video understanding
show that DynFrame-4B is competitive with strong 7B--8B baselines,
while DynFrame-8B achieves new best results on most metrics.

%% file: appendix.tex
\clearpage
\appendix
\section*{Supplementary Material}
\setlength{\textfloatsep}{4pt plus 1pt minus 1pt}
\setlength{\floatsep}{4pt plus 1pt minus 1pt}
\setlength{\intextsep}{4pt plus 1pt minus 1pt}
\setlength{\abovecaptionskip}{3pt}
\setlength{\belowcaptionskip}{0pt}
\setlength{\parskip}{0.2em}
\setlength{\dbltextfloatsep}{5pt plus 1pt minus 1pt}
\setlength{\dblfloatsep}{5pt plus 1pt minus 1pt}
\renewcommand{\textfraction}{0.03}
\renewcommand{\floatpagefraction}{0.9}
\renewcommand{\dblfloatpagefraction}{0.9}
\renewcommand{\topfraction}{0.97}
\renewcommand{\dbltopfraction}{0.97}
\setcounter{section}{0}
\renewcommand{\thesection}{\Alph{section}}
\setcounter{figure}{0}
\renewcommand{\thefigure}{A\arabic{figure}}
\setcounter{table}{0}
\renewcommand{\thetable}{A\arabic{table}}
\setcounter{equation}{0}
\renewcommand{\theequation}{A\arabic{equation}}

\section{Data Generation Prompts}
\label{sec:appendix_prompts}

We provide the complete prompts used to query
Gemini-3-Pro~\cite{comanici2025gemini25} for constructing our
training data. We design two complementary prompts tailored to
the annotation characteristics of each task type. For VideoQA
and Grounded VQA, where only question--answer pairs are
available without temporal annotations, we query Gemini to
produce both temporal localization and answer reasoning from
scratch. For temporal grounding, where ground-truth temporal
boundaries already exist, we adopt a reformulative strategy by
prompting Gemini to expand, clean, and canonicalize existing
reasoning traces while preserving the original annotations.
Both prompts share a unified adaptive FPS selection guideline to
ensure consistent, content-aware sampling across the training
mixture.

\subsection{VideoQA \& Grounded VQA Prompt}
\label{sec:appendix_prompt_vqa}

The prompt in Figure~\ref{fig:prompt_vqa} performs two-stage
annotation in a single API call via a structured JSON with two
fields. In \texttt{zoom\_in\_cot}, the model reasons about which
temporal segment contains the required visual evidence and
concludes with \texttt{<time\_span>} and \texttt{<fps>} tags,
without predicting or guessing the answer. In
\texttt{answer\_cot}, the model assumes it has watched
\textbf{only} that segment at the specified FPS and reasons step
by step to produce the final answer. A three-tier FPS guideline
(1--2\,fps for static scenes, 3--4\,fps for moderate dynamics,
5--6\,fps for rapid actions) is embedded to ensure
activity-adaptive sampling density.

\subsection{Temporal Grounding Prompt}
\label{sec:appendix_prompt_tg}

The prompt in Figure~\ref{fig:prompt_tg} reformulates existing
reasoning traces rather than annotating from scratch. Given a
trace with ground-truth temporal boundaries, the model expands
the span by a random margin of $0.5$--$2.0$\,s on each side for
contextual retrieval, relocates the expanded
\texttt{<time\_span>} and an adaptive \texttt{<fps>} tag to the
front of the chain, and removes redundant temporal descriptions.
The original unexpanded boundaries serve as the supervision
signal.

\begin{figure*}[!t]
\begin{tcolorbox}[
  colback=gray!10,
  colframe=gray!60!black,
  title=\textbf{Prompt for VideoQA \& Grounded VQA Data Generation},
  fonttitle=\bfseries,
  arc=2mm,
  boxrule=1pt,
]
\scriptsize

\noindent\textbf{System Prompt:} You are a video analysis assistant. Answer based solely on the visual content of the video. Rely strictly on visual cues; ignore audio. Respond in valid, raw JSON only.

\par\smallskip
\noindent\textcolor{gray!70}{\dotfill}
\par\smallskip

\noindent\textbf{User Prompt:}

You will respond in valid, raw JSON with exactly two top-level keys.

\par\smallskip
\noindent\textbf{JSON Structure:}
\begin{verbatim}
{
  "zoom_in_cot": "Your step-by-step reasoning for selecting
   the specific video segment. This string MUST end strictly
   with the time span and FPS tags in the format
   <span>START - END</span> <fps>N</fps>.",

  "answer_cot": "Your step-by-step reasoning after 'watching'
   the selected segment. This string MUST end strictly with
   the final answer wrapped in
   <answer>OPTION_LETTER</answer>."
}
\end{verbatim}

\noindent\textbf{Workflow}
\begin{enumerate}[leftmargin=*, nosep]
    \item Read the user's question and infer the precise visual evidence required.
    \item Describe, step by step, how you would locate the segment that likely contains this evidence. Explain why this portion is critical.
    \item Based on the nature and speed of the target activity, recommend an appropriate FPS for analysis.
    \item Conclude your ``zoom\_in\_cot'' reasoning by appending the time span and FPS tags. \textbf{Do NOT} provide or guess the answer in this field.
    \item After that, imagine you have now watched \textbf{ONLY} that segment at the specified FPS. Using the visual information in the segment, reason step-by-step to reach the answer.
    \item This reasoning plus the final answer (wrapped in \texttt{<answer></answer>}) goes into ``answer\_cot''.
\end{enumerate}

\noindent\textbf{FPS Selection Guideline}
\begin{itemize}[leftmargin=*, nosep]
    \item \textbf{1--2\,fps}: Static or quasi-static scenes with minimal temporal variation (e.g., reading text/signs, appearance attributes, object identification, colors, a person standing still).
    \item \textbf{3--4\,fps}: Moderately dynamic scenarios with clear temporal progression (e.g., general activities, object interactions, walking, cooking step-by-step, gesture recognition, conversations).
    \item \textbf{5--6\,fps}: Rapid, fine-grained actions requiring frame-by-frame analysis (e.g., fast actions, rapid motion, counting quick events, sports movements, catching/throwing a ball).
\end{itemize}

\noindent\textbf{Strict Rules}
\begin{itemize}[leftmargin=*, nosep]
    \item Output must be raw JSON only (no markdown fences like \texttt{```json}).
    \item Time span must use seconds with two decimal places (e.g., 5.00 - 9.00).
    \item The ``zoom\_in\_cot'' value must end with the closing tag \texttt{</fps>}. Do not add any punctuation, whitespace, or words after this tag.
    \item The final answer must appear only once, inside \texttt{<answer></answer>}, at the very end of ``answer\_cot''.
    \item Inside \texttt{<answer></answer>}, output only the letter A, B, C, D\ldots{} (do not include the option text).
    \item FPS must be an integer between 1 and 6.
\end{itemize}

\vspace{2mm}
\noindent\rule{\linewidth}{0.4pt}
\textbf{Question:} \{question\} \\
\textbf{Options:} \{options\}

\end{tcolorbox}
\caption{\textbf{Data generation prompt for VideoQA.} We prompt
Gemini-3-Pro to jointly perform temporal evidence
identification with adaptive FPS recommendation and
clip-constrained answer generation in a single call via two
structured JSON fields. The FPS selection guideline ensures the
sampling rate matches the visual dynamics of the target
activity.}
\label{fig:prompt_vqa}
\end{figure*}

\clearpage

\begin{figure*}[p]
\begin{tcolorbox}[
  colback=gray!10,
  colframe=gray!60!black,
  title=\textbf{Prompt for Temporal Grounding Data Generation},
  fonttitle=\bfseries,
  arc=2mm,
  boxrule=1pt,
]
\scriptsize

\noindent\textbf{System Prompt:} You are a video analysis assistant specializing in temporal grounding. Respond in valid, raw text only following the exact output format specified below.

\par\smallskip
\noindent\textcolor{gray!70}{\dotfill}
\par\smallskip

\noindent\textbf{User Prompt:}

You are given a temporal grounding reasoning trace that contains a thinking process and an answer. Your task is to \textbf{reformulate} it into our canonical format, \textbf{expand the temporal window for contextual evidence}, and \textbf{recommend an appropriate FPS} for the identified temporal span. Perform the following operations:

\par\smallskip
\noindent\textbf{Workflow}
\begin{enumerate}[leftmargin=*, nosep]
    \item Starting from the original temporal span in the input, expand the start and end boundaries by a random margin of 0.5--2.0 seconds on each side to provide additional temporal context.
    \item Move the expanded temporal span to the front of the thinking process, formatted as \texttt{<span>START - END</span>} (seconds with two decimal places, e.g., \texttt{<span>29.50 - 74.00</span>}).
    \item Remove redundant or repeated temporal descriptions in the thinking process. Keep the reasoning concise and non-repetitive.
    \item Based on the duration of the \textbf{expanded span} and the nature of the activity described, append an FPS tag \texttt{<fps>N</fps>} immediately after the \texttt{<span>} tag.
    \item Extract only the \textbf{original} start and end timestamps as the final answer (e.g., \texttt{30.00 - 72.00}).
\end{enumerate}

\par\smallskip
\noindent\textbf{FPS Selection Guideline}
\begin{itemize}[leftmargin=*, nosep]
    \item \textbf{1--2\,fps}: Static or quasi-static scenes with minimal temporal variation (e.g., reading text/signs, appearance attributes, object identification, colors, a person standing still).
    \item \textbf{3--4\,fps}: Moderately dynamic scenarios with clear temporal progression (e.g., general activities, object interactions, walking, cooking step-by-step, gesture recognition, conversations).
    \item \textbf{5--6\,fps}: Rapid, fine-grained actions requiring frame-by-frame analysis (e.g., fast actions, rapid motion, counting quick events, sports movements, catching/throwing a ball).
\end{itemize}

\par\smallskip
\noindent\textbf{Output Format}
\begin{verbatim}
<think>
<span>EXPANDED_START - EXPANDED_END</span> <fps>N</fps>
[Rest of the cleaned reasoning...]
</think>
<answer>ORIGINAL_START - ORIGINAL_END</answer>
\end{verbatim}

\par\smallskip
\noindent\textbf{Strict Rules}
\begin{itemize}[leftmargin=*, nosep]
    \item Inside \texttt{<think>}, the first line must be \texttt{<time\_span>EXPANDED\_START - EXPANDED\_END</time\_span> <fps>N</fps>}.
    \item FPS must be an integer between 1 and 6.
    \item The \texttt{<time\_span>} must contain the expanded temporal span.
    \item The \texttt{<answer>} must contain only the original, unexpanded start and end timestamps, with no additional text.
    \item Remove redundant temporal descriptions but preserve the core reasoning logic.
\end{itemize}

\vspace{2mm}
\noindent\rule{\linewidth}{0.4pt}
\textbf{Input:} \{current\_input\}\\
\textbf{Output:}

\end{tcolorbox}
\caption{\textbf{Data reformulation and FPS annotation prompt
for temporal grounding.} Given an existing reasoning trace with
ground-truth temporal boundaries, we prompt Gemini-3-Pro to
canonicalize the format by expanding the temporal window for
contextual evidence, relocating the contextual span to the
front, cleaning redundant descriptions, recommending an
activity-adaptive FPS, and extracting the original minimal
answer.}
\label{fig:prompt_tg}
\end{figure*}

\clearpage

\setcounter{figure}{0}
\renewcommand{\thefigure}{B\arabic{figure}}

\section{Analysis of Temporal Span and FPS Prediction}
\label{app:span_fps_analysis}

We analyze the temporal span and FPS predictions of DynFrame
to illustrate its learned retrieval behavior.

\subsection{Temporal Span Prediction}
\label{app:span_pred}

\begin{figure*}[!t]
    \centering
    \begin{subfigure}[t]{0.48\linewidth}
        \centering
        \includegraphics[width=\linewidth]{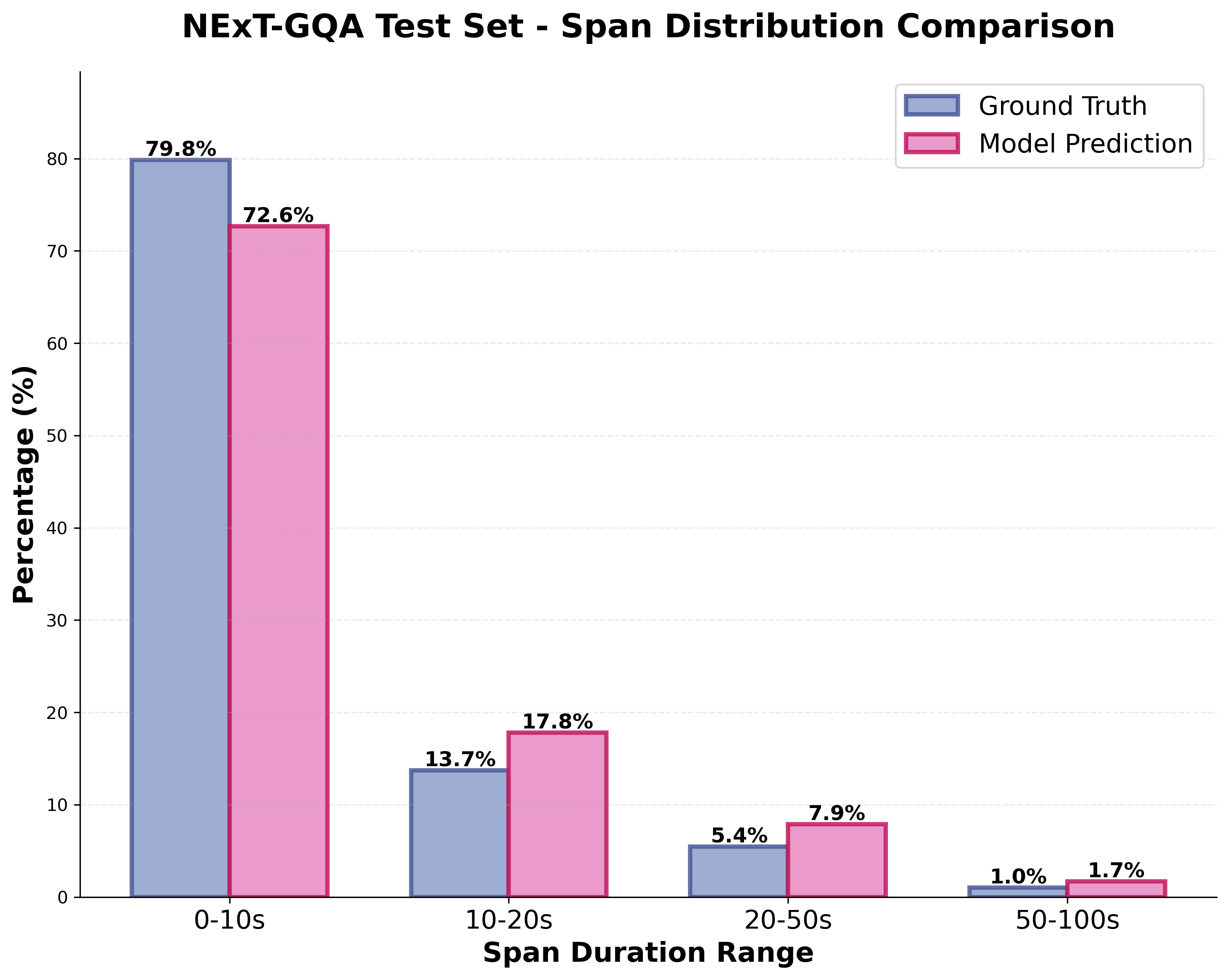}
        \caption{NExT-GQA test set.}
        \label{fig:span_nextgqa}
    \end{subfigure}\hfill%
    \begin{subfigure}[t]{0.48\linewidth}
        \centering
        \includegraphics[width=\linewidth]{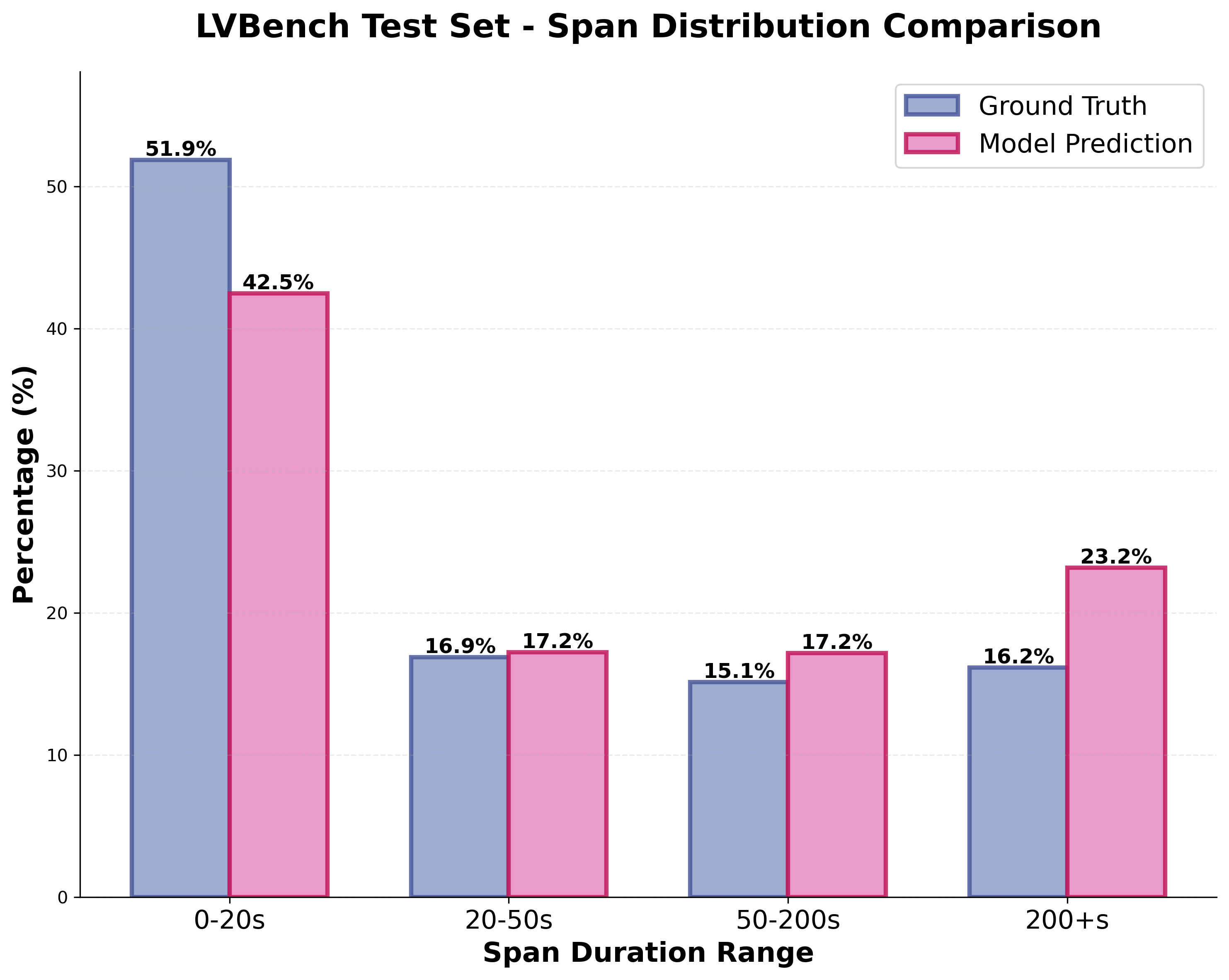}
        \caption{LVBench test set.}
        \label{fig:span_lvbench}
    \end{subfigure}
    \caption{\textbf{Temporal span distribution comparison}
    between ground-truth annotations and model predictions.
    (a)~NExT-GQA; (b)~LVBench. Across both benchmarks, the model
    systematically shifts probability mass from shorter to
    longer spans, predicting broader temporal windows to capture
    sufficient contextual evidence for reasoning.}
    \label{fig:span_comparison}
\end{figure*}

Figure~\ref{fig:span_comparison} compares the predicted and
ground-truth span distributions on NExT-GQA (short-video) and
LVBench (long-video). On NExT-GQA, ground-truth spans are
heavily concentrated in the $0$--$10$\,s range, and the model's
predictions closely follow this distribution while slightly
redistributing mass toward longer spans, indicating accurate
localization on short-form videos. On LVBench, where
ground-truth spans are distributed more evenly across all
duration ranges and are notably longer overall, the model
consistently predicts correspondingly broader temporal windows
for extended segments---mirroring the same adaptive behavior
observed on shorter videos. This consistency across vastly
different video lengths confirms that the model has learned a
robust, duration-aware grounding strategy rather than a
dataset-specific bias, with predictions shifting moderately
toward wider windows to capture sufficient contextual evidence
for downstream reasoning.

\subsection{Dynamic FPS Prediction}
\label{app:fps_pred}

\begin{figure}[!htbp]
    \centering
    \includegraphics[width=0.5\linewidth]{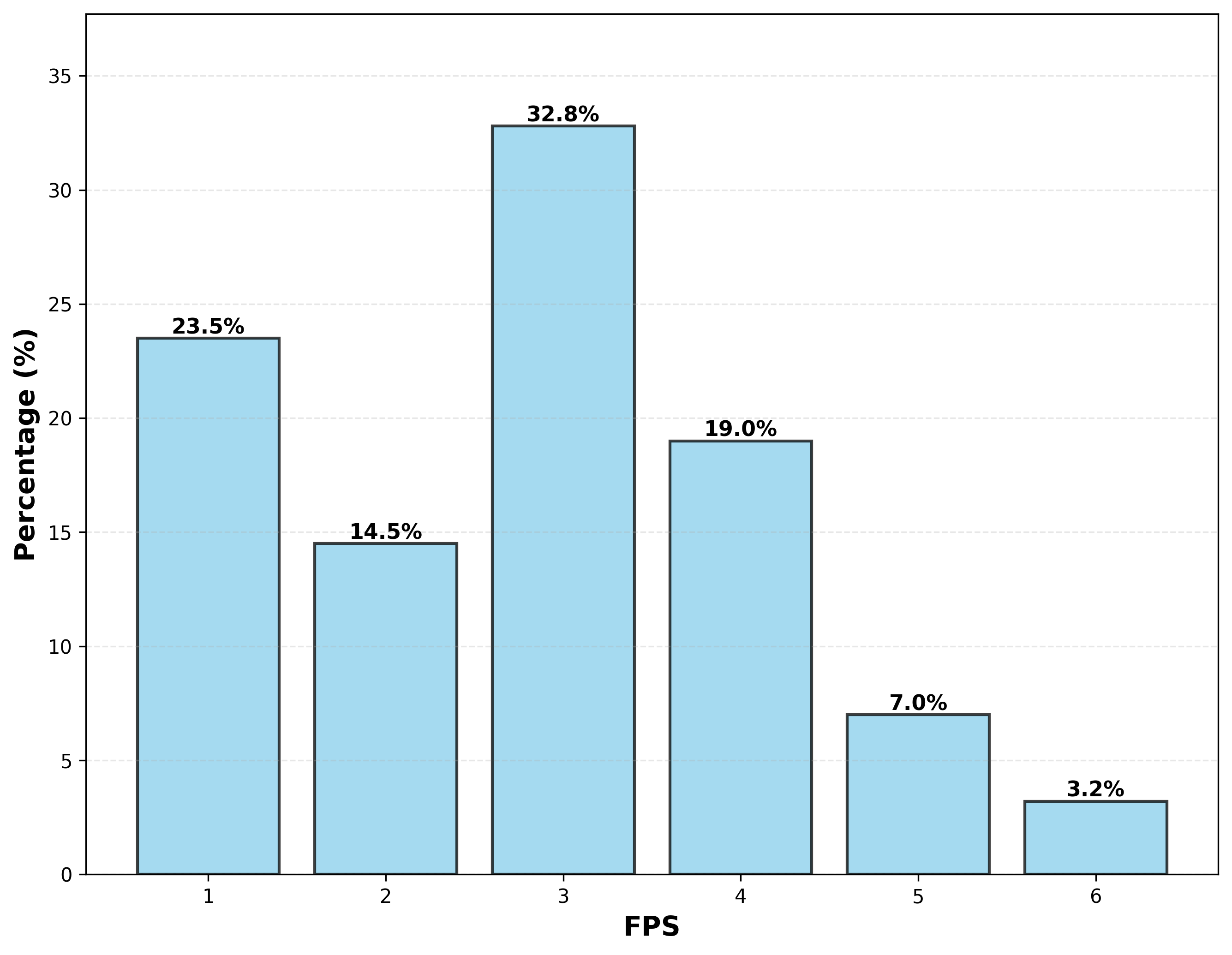}
    \caption{\textbf{Predicted retrieval FPS distribution
    ($f_2$) of DynFrame.}}
    \label{fig:fps_dist}
\end{figure}

Figure~\ref{fig:fps_dist} shows the distribution of predicted
retrieval FPS ($f_2$) of DynFrame. The majority of predictions
fall within $1$--$4$\,fps. This concentration is well-aligned
with the nature of most video understanding tasks, where
moderate frame rates already provide sufficient visual
information for accurate reasoning. Nevertheless, for highly
dynamic scenes that demand dense temporal sampling to capture
rapid changes within very short intervals, DynFrame also
correctly predicts higher frame rates ($5$--$6$\,fps),
demonstrating its ability to adapt sampling density to content
complexity.

\begin{figure}[!htbp]
    \centering
    \begin{subfigure}[t]{0.32\linewidth}
        \centering
        \includegraphics[width=\linewidth]{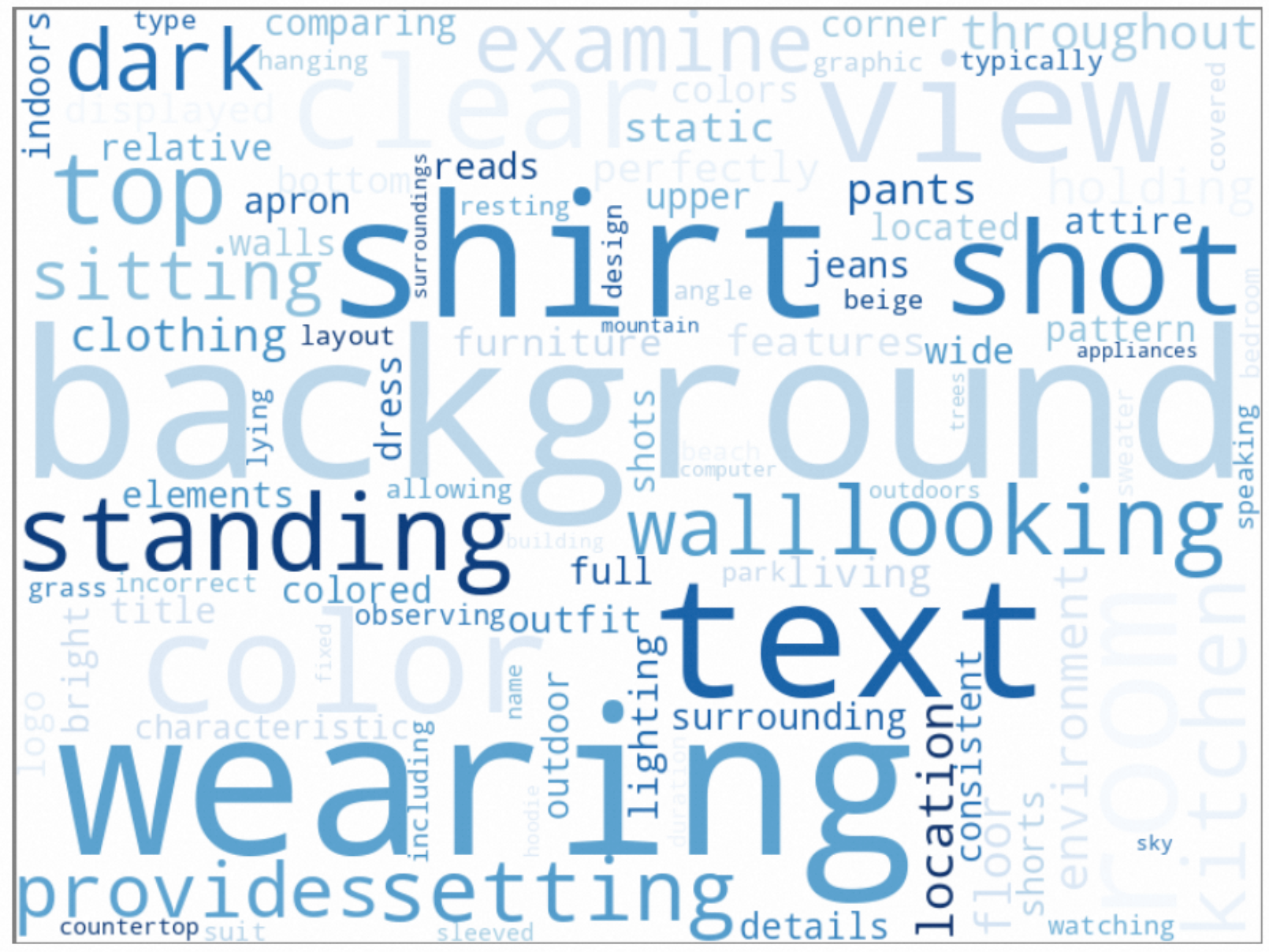}
        \caption{1--2\,FPS}
        \label{fig:fps_wc_low}
    \end{subfigure}\hfill%
    \begin{subfigure}[t]{0.32\linewidth}
        \centering
        \includegraphics[width=\linewidth]{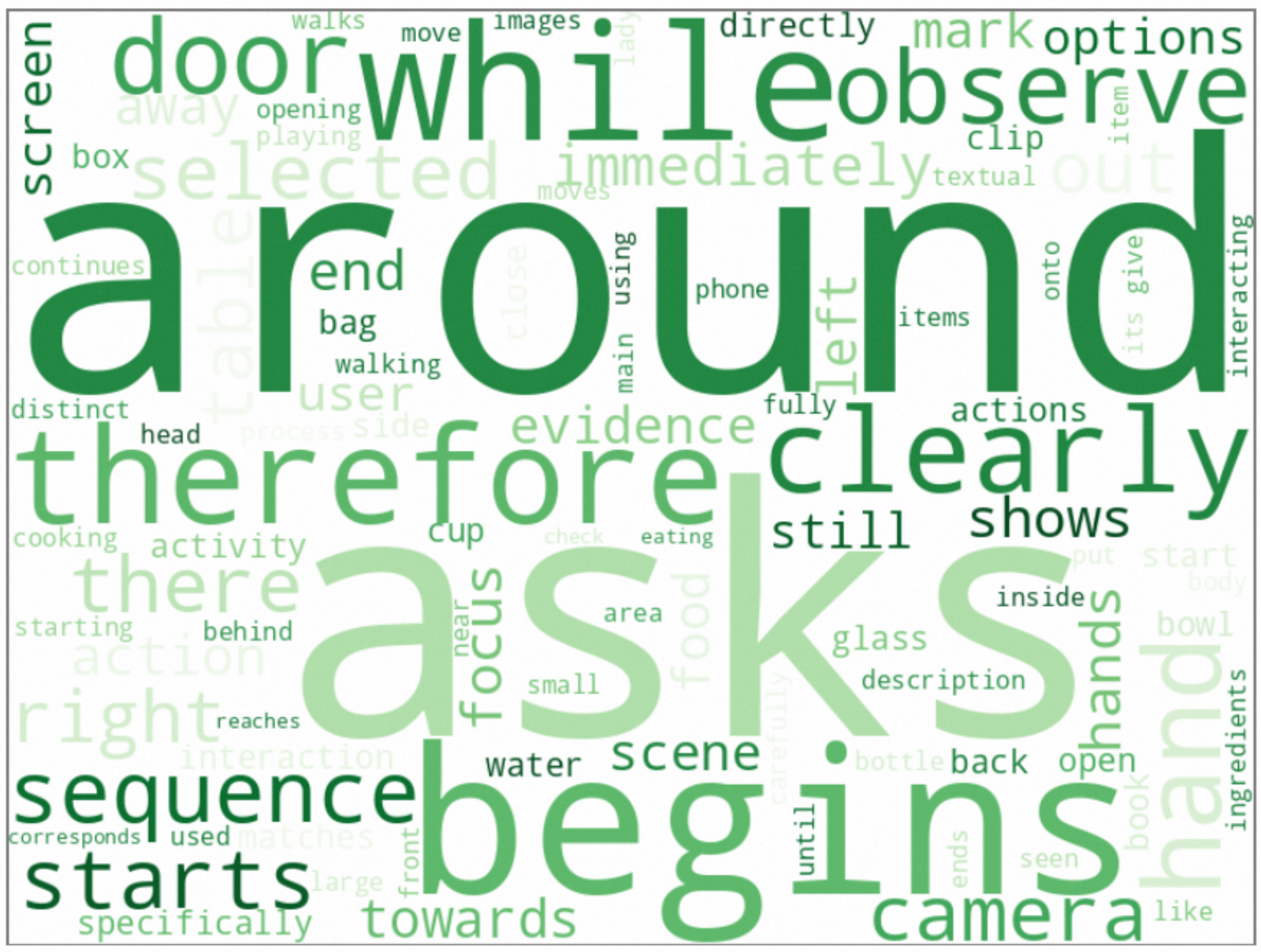}
        \caption{3--4\,FPS}
        \label{fig:fps_wc_mid}
    \end{subfigure}\hfill%
    \begin{subfigure}[t]{0.32\linewidth}
        \centering
        \includegraphics[width=\linewidth]{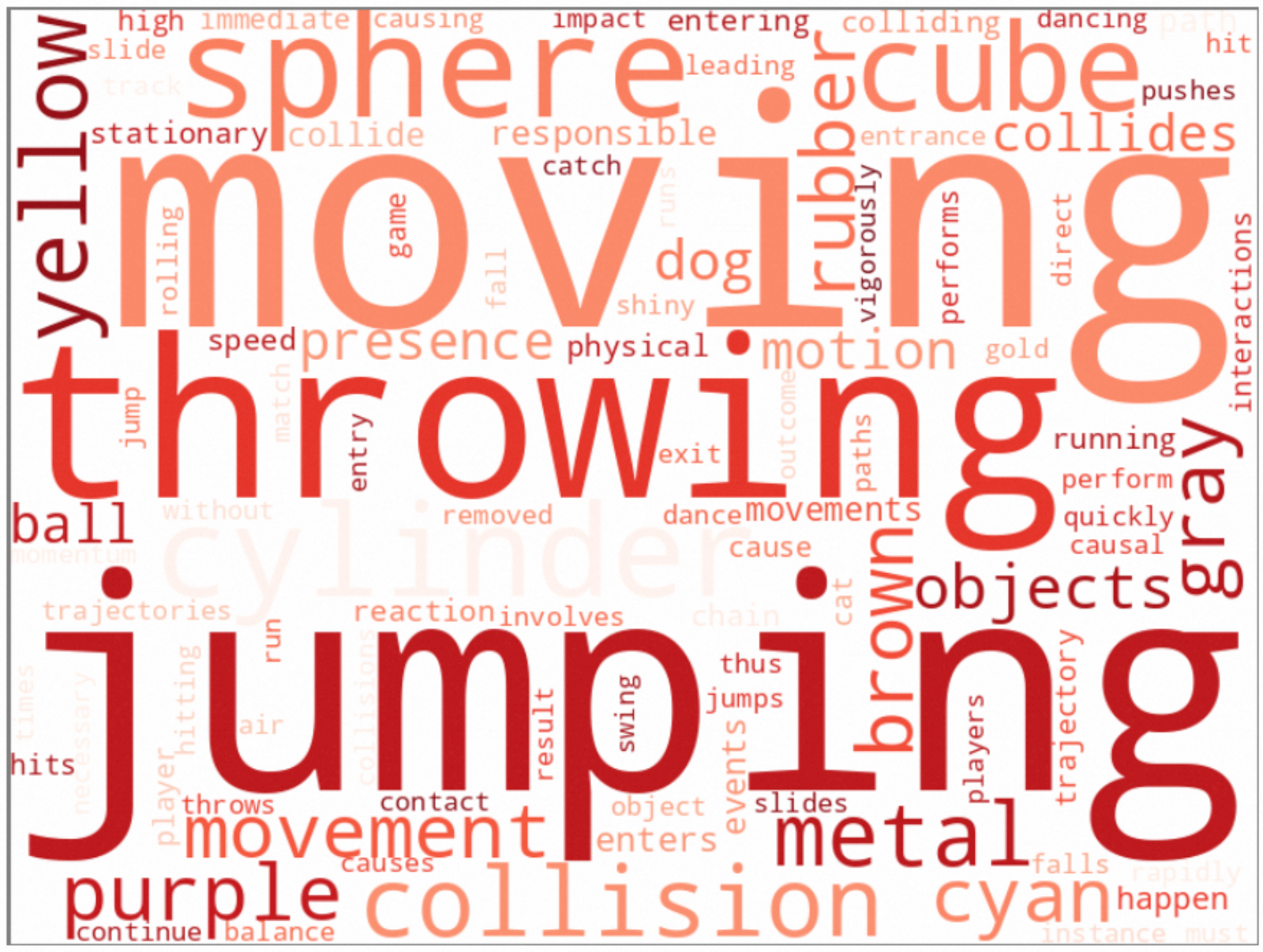}
        \caption{5--6\,FPS}
        \label{fig:fps_wc_high}
    \end{subfigure}
    \caption{\textbf{Word clouds of content keywords by
    retrieval FPS band.} Low FPS captures static descriptors;
    medium FPS captures sequential activities; high FPS
    captures rapid actions.}
    \label{fig:fps_wordcloud}
\end{figure}

To further investigate what drives different FPS predictions,
we extract the most frequent content keywords within three FPS
bands and visualize them as deduplicated word clouds
(Figure~\ref{fig:fps_wordcloud}). A clear semantic gradient
emerges across the bands. \textbf{Low FPS ($1$--$2$\,fps)} is
dominated by static descriptors: \textit{standing},
\textit{wearing}, \textit{background}, \textit{shirt},
\textit{wall}, \textit{text}---scene descriptions and appearance
attributes with minimal temporal variation. \textbf{Medium FPS
($3$--$4$\,fps)} features sequential-activity terms:
\textit{around}, \textit{begins}, \textit{sequence},
\textit{asks}, \textit{observe}---multi-step procedures and
conversational interactions that require tracking temporal
progression but not rapid motion. \textbf{High FPS
($5$--$6$\,fps)} is characterized by rapid-action keywords:
\textit{moving}, \textit{throwing}, \textit{jumping},
\textit{collision}, \textit{sphere}, \textit{cube}---fast-paced
physical interactions where dense sampling is essential to
capture critical state transitions. This semantic stratification
confirms that DynFrame learns a meaningful mapping from
content dynamics to sampling frequency. Together with the span
prediction strategy analyzed above, the two mechanisms work
synergistically: broader spans increase temporal coverage,
while adaptive FPS controls sampling density within the
retrieved window.

\setcounter{figure}{0}
\renewcommand{\thefigure}{C\arabic{figure}}
\setcounter{table}{0}
\renewcommand{\thetable}{C\arabic{table}}

\section{Frame-Budget and Context-Length Protocol}
\label{app:cost_protocol}

Table~\ref{tab:context_protocol} reports the inference context
protocol used by recent thinking-with-video methods and our
\textbf{DynFrame}. We compare
the average single-forward context length for each benchmark.
Multi-round methods may additionally incur repeated forward calls and
larger cumulative token costs.

\begin{table*}[!htb]
\centering
\renewcommand{\arraystretch}{1.12}
\caption{
\textbf{Average single-forward context length across benchmarks.}
Numbers are measured in tokens. ``Max retrieval / injection'' denotes
the maximum number of visual revisiting operations used by each method:
tool calls, zoom-in calls, iterative perception rounds, or visual-token
injections. ``0'' indicates no visual retrieval after the initial input.
}
\label{tab:context_protocol}
\resizebox{\textwidth}{!}{%
\begin{tabular}{l|c|cccccc}
\toprule[1pt]
\multirow{2}{*}{\textbf{Model}}
& \multirow{2}{*}{\textbf{Max retrieval / injection}}
& \multicolumn{6}{c}{\textbf{Average single-forward context length}} \\
\cmidrule(lr){3-8}
& & \textbf{NExT-GQA} & \textbf{Charades} & \textbf{ActivityNet}
& \textbf{Video-MME} & \textbf{MLVU} & \textbf{LVBench} \\
\midrule\midrule

Qwen2.5-VL-7B~\cite{bai2025qwen2.5vl}
& 0
& 2910.9 & 1748.0 & 6282.4 & 2021.8 & 1991.5 & 2021.3 \\

Video-R1-7B~\cite{feng2025videor1}
& 0
& 3893.2 & 3009.1 & 5482.5 & 4238.9 & 3968.4 & 4214.3 \\

VideoZoomer~\cite{ding2025videozoomer}
& 4
& 12254.9 & 11195.1 & 11047.0 & 15745.0 & 17518.3 & 17896.7 \\

Video-o3~\cite{zeng2026videoo3}
& 8
& 55136.3 & 45716.0 & 44341.6 & 59134.0 & 61140.0 & 68478.1 \\

VideoChat-R1.5-7B~\cite{videochatr152025}
& 2
& 8897.5 & 5204.6 & 8213.9 & 38485.8 & 49872.9 & 46683.1 \\

LongVT-RFT~\cite{yang2025longvt}
& 5
& 16389.0 & 10398.7 & 24113.6 & 33278.1 & 49606.9 & 55497.7 \\

LOVE-R1-stage3~\cite{lover12025}
& 2
& 8480.5 & 4662.3 & 5757.5 & 10992.1 & 16919.5 & 18859.9 \\

\midrule
\rowcolor{gray!18}
\textbf{DynFrame-8B (ours)}
& 1
& 6734.5 & 4936.7 & 5682.7 & 8253.7 & 7964.2 & 9749.6 \\

\bottomrule[1pt]
\end{tabular}
}
\end{table*}

\paragraph{Analysis.}
Table~\ref{tab:context_protocol} shows that methods without visual
retrieval have small and stable maximum contexts. Qwen2.5-VL-7B and
Video-R1-7B require only one forward pass. For VideoZoomer and Video-o3, the official repositories enable
multi-round visual revisiting: VideoZoomer uses up to four
\texttt{<video\_zoom>} calls, while Video-o3 uses up to eight
\texttt{<grounding>} observations in its main multi-turn video-QA
protocol. The corresponding rows are measured with these maximum
retrieval trajectories rather than with single-turn Qwen-style context
lengths.

In contrast, multi-round thinking-with-video systems substantially
increase the average context length once retrieved visual evidence is
inserted into the inference trajectory. VideoChat-R1.5, which uses
iterative perception, reaches much larger contexts on long-form
benchmarks such as Video-MME, MLVU, and LVBench. Tool-enabled
LongVT-RFT further increases the maximum context length, especially on
long-video tasks, because up to five retrieved clips can be appended
across multiple rounds. LOVE-R1 is lighter than LongVT-RFT-tool but
still produces noticeably larger contexts than single-forward baselines
due to its fast-view plus zoom-in reasoning design.

DynFrame follows a different design point. It uses only one
model-predicted visual injection to jointly specify the temporal window
and the sampling density. This makes each visual revisit more
task-adaptive than fixed-policy retrieval calls. This design not only reduces the
reliance on repeated retrieval rounds, but also improves accuracy over
recent retrieval-based thinking-with-video methods on most evaluated
benchmarks.

\setcounter{figure}{0}
\renewcommand{\thefigure}{D\arabic{figure}}

\section{Additional Qualitative Examples}
\label{app:qualitative_examples}

We present additional qualitative examples to illustrate the
behavior of DynFrame across diverse question types and video
domains. Each example shows the model's full reasoning
trajectory, including the predicted temporal span, adaptive FPS
selection, and the grounded answer derivation.

\begin{figure}[!htb]
    \centering
    \includegraphics[width=0.95\linewidth]{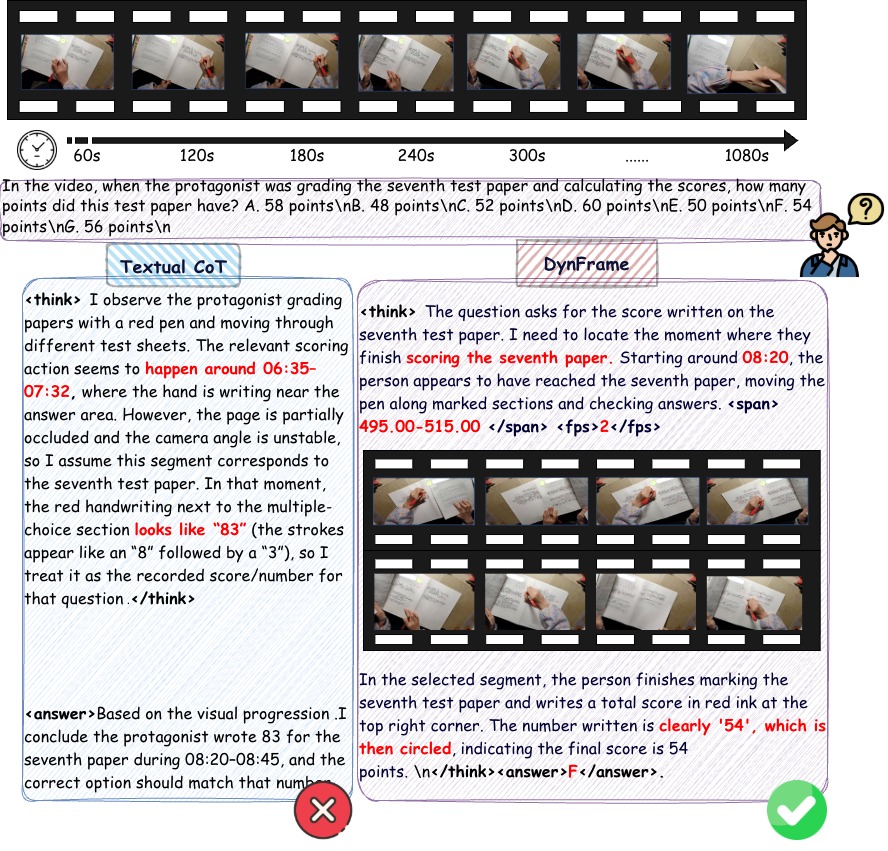}
    \caption{\textbf{Case study 1: long-video document
    counting.} DynFrame correctly identifies the 7th exam
    paper in the long video and accurately recognizes the
    scores in both grading rounds, whereas the textual CoT
    baseline fails.}
    \label{fig:case_study_1}
\end{figure}

\begin{figure}[!htb]
    \centering
    \includegraphics[width=0.95\linewidth]{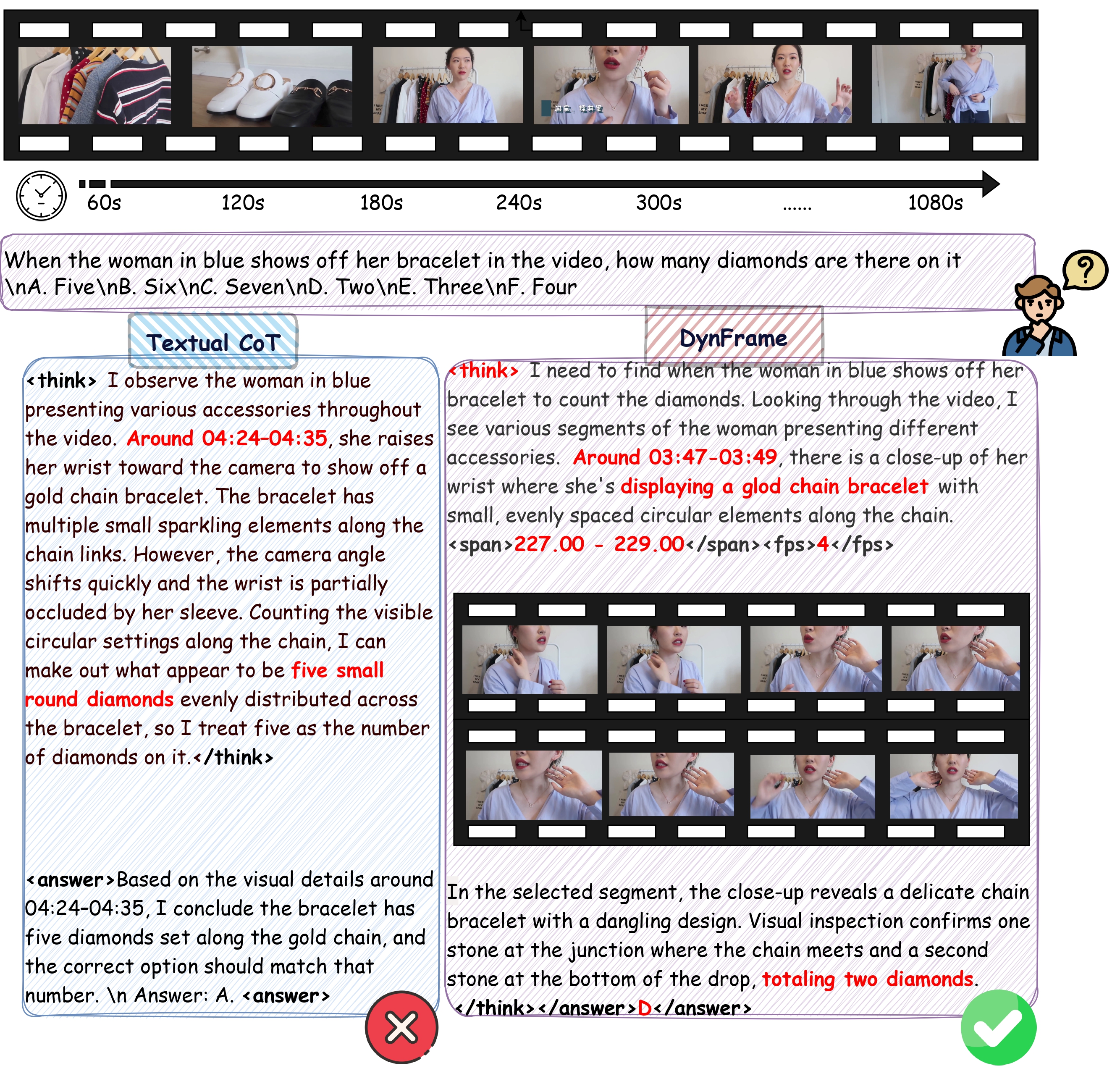}
    \caption{\textbf{Case study 2: fine-grained object
    counting.} DynFrame successfully locates the $2$\,s clip
    showcasing the bracelet in the long video and accurately
    counts the number of diamonds at $4$\,FPS, whereas the
    textual CoT baseline fails.}
    \label{fig:case_study_2}
\end{figure}